\documentclass{article}

\usepackage{microtype}
\usepackage{graphicx}
\usepackage{subfigure}
\usepackage{booktabs} 
\usepackage{multirow}  

 \usepackage[accepted]{icml2025}

\usepackage{amsmath}
\usepackage{amssymb}
\usepackage{mathtools}
\usepackage{amsthm}

\usepackage[capitalize,noabbrev]{cleveref}

\theoremstyle{plain}

\theoremstyle{definition}

\theoremstyle{remark}

\usepackage[textsize=tiny]{todonotes}
\usepackage{subcaption}

\begin{document}

\twocolumn[
\icmltitle{Label-Efficient Hyperspectral Image Classification via\\
Spectral FiLM Modulation of Low-Level Pretrained Diffusion Features}




\begin{icmlauthorlist}

\icmlauthor{Yuzhen Hu}{uh}
\icmlauthor{Biplab Banerjee}{iitb}
\icmlauthor{Saurabh Prasad}{uh}
\end{icmlauthorlist}

\icmlaffiliation{uh}{University of Houston, Texas, USA}
\icmlaffiliation{iitb}{Indian Institute of Technology Bombay, Mumbai, India}


\icmlcorrespondingauthor{Yuzhen Hu}{yhu34@uh.edu}
\icmlcorrespondingauthor{Saurabh Prasad}{saurabh.prasad@ieee.org}


\icmlkeywords{Diffusion Models, Hyperspectral Imaging, Remote Sensing, Label-Efficient Learning, Land Cover Mapping}

\vskip 0.3in
]



\printAffiliationsAndNotice{}
\begin{abstract}
Hyperspectral imaging (HSI) enables detailed land cover classification, but low spatial resolution and sparse annotations pose significant challenges. We present a label-efficient framework that leverages spatial features from a frozen diffusion model pretrained on natural images. Specifically, we extract low-level representations from high-resolution decoder layers at early denoising timesteps, which transfer well to the low-texture setting of HSI. To combine spectral and spatial information, we introduce a lightweight FiLM-based fusion module that adaptively integrates spectral cues into frozen spatial features, enabling effective multimodal learning under sparse supervision. Experiments on two recent hyperspectral datasets show that our method outperforms state-of-the-art approaches using only the sparse training labels provided. Ablation studies further validate the benefit of diffusion-based features and spectral-aware fusion. Our results suggest that pretrained diffusion models can support domain-agnostic, label-efficient representation learning in remote sensing and scientific imaging tasks.
\end{abstract}

\section{Introduction}
\label{sec:intro}

Land cover mapping is a fundamental task in remote sensing, supporting applications such as environmental monitoring, agriculture, and resource management. 
Hyperspectral images (HSIs), with their dense spectral reflectance information, provide detailed insights into material properties and are well-suited for this purpose.

Despite their rich spectral content, HSIs pose several challenges. Their high dimensionality increases computational cost and overfitting risk, especially under limited supervision. 
Moreover, the trade-off between spectral fidelity and spatial resolution often leads to poor spatial detail, limiting segmentation accuracy.

Additional challenges arise from the spectral and spatial variability of land cover types across regions. 
Subtle intra-class variations—due to differences in vegetation, soil, or human activity—make generalization difficult~\cite{akiva2022self,liu2024rotated,prasad2024advances,kumar2023methanemapper,li2023spectral}. 
Finally, acquiring high-quality labeled data is expensive and time-consuming, as pixel-level annotation requires domain expertise. These limitations highlight the need for approaches that can extract robust features without heavy reliance on labels. 
Unsupervised and self-supervised learning methods address this by learning directly from the data.

Generative models—particularly diffusion models~\cite{ho2020denoising,song2020score,sohl2015deep,dhariwal2021diffusion,pang2024hir}—have recently shown state-of-the-art performance in image synthesis, restoration, and manipulation tasks. 
Through unsupervised training and iterative denoising, diffusion models learn the underlying data distribution, enabling them to capture rich spatial structures and pixel-wise contextual dependencies—traits especially valuable for segmentation and representation learning under data scarcity.

Unlike deterministic self-supervised methods such as masked autoencoders~\cite{he2022masked}, diffusion models operate in a probabilistic framework that better handles uncertainty and degraded inputs. 
This makes them well-suited for low-resolution, low-texture hyperspectral imagery~\cite{chen2020generative,zhang2023tale}. 
Notably, \cite{baranchuk2021label} demonstrated that pre-trained diffusion models can provide strong pixel-level representations, outperforming earlier self-supervised methods under limited supervision and maintaining robustness under corrupted inputs.
Pre-trained diffusion models have shown impressive performance in natural image domains~\cite{xu2023open,zhang2023tale}, yet their application to geospatial imagery remains underexplored. This is due in part to significant domain shifts—differences in spatial scale, viewing geometry, and spectral coverage (e.g., near and short-wave infrared)—which challenge cross-domain generalization~\cite{wang2021pretraining,zhang2023adding}.

Beyond domain shift, hyperspectral imagery introduces a unique modality challenge: each pixel contains a high-dimensional spectral signature critical for land cover analysis. Effectively leveraging both spectral and spatial information—especially under limited supervision—requires models capable of adaptive fusion conditioned on spectral features. To address this, we adopt FiLM-based modulation ~\cite{perez2018film} , a lightweight and parameter-efficient conditioning method that remains largely unexplored in hyperspectral settings.

This work addresses two key challenges in hyperspectral land cover mapping: (1) transferring pre-trained diffusion models to geospatial domains, and (2) enabling adaptive fusion of spatial and spectral modalities under sparse supervision. We introduce \textbf{GeoDiffNet-F}, a label-efficient framework that reuses a diffusion model pre-trained on natural images to extract transferable spatial features, which are fused with spectral embeddings using FiLM-based modulation.
Our contributions are as follows:\\
(1) We propose \textbf{GeoDiffNet}, a lightweight framework that repurposes frozen decoder layers of pre-trained diffusion models to extract low-level spatial features that generalize well to geospatial imagery with weak texture and low resolution;\\
(2) we demonstrate strong cross-domain transferability of diffusion features without requiring domain-specific finetuning;\\
(3) we introduce \textbf{GeoDiffNet-F}, which fuses spectral and spatial features using FiLM-based modulation, enabling dynamic, feature-wise conditioning from spectral input;\\
(4) we perform a detailed transferability analysis across decoder layers and denoising timesteps, showing that early diffusion features are robust to significant domain shift.

\section{Related work}
\label{sec:relate_work}
\textbf{Feature Transferability in Deep Learning}
 has been extensively studied, particularly with convolutional neural networks (CNNs). Early layers in CNNs capture low-level features, such as edges and textures, which are highly transferable across different tasks and datasets \cite{yosinski2014transferable,long2015unsupervised}. This principle has been foundational in the success of transfer learning, enabling models pre-trained on large-scale datasets to be fine-tuned for specific tasks with smaller datasets.
 
\textbf{Diffusion Models} as a new class of generative models, have shown remarkable performance in generating high-fidelity images. These models learn to generate data by reversing a diffusion process, progressively transforming noise into structured data. Recent advancements \cite{ho2020denoising, song2020score, dhariwal2021diffusion} have positioned diffusion models as state-of-the-art in image generation tasks.

\textbf{Diffusion Models for Feature Extraction.}
The potential of diffusion models for feature extraction has attracted growing interest. Prior studies \cite{baranchuk2021label, xu2023open, zhang2023tale, luo2023diffusion} extract features from various layers of the U-Net architecture and timesteps in the diffusion process, leveraging multi-scale and multi-timestep information for robust pixel-level descriptors in tasks like image segmentation. However, these works assume a well-aligned source and target domain. To date, no studies have investigated the use of diffusion-based features in cross-domain settings, leaving their generalizability underexplored.

\textbf{Diffusion Models in Remote Sensing.}
Diffusion models have recently gained attention in remote sensing \cite{bandara2022ddpm, ayala2023diffusion, bai2022conditional, chen2023spectraldiff, pang2024hir}. Most prior work trains models from scratch or tailors them to specific datasets—e.g., \cite{chen2023spectraldiff} proposes a 3D diffusion model for hyperspectral data that demands substantial computation and large training sets. In contrast, we are the first to evaluate a universal pre-trained diffusion model  for geospatial analysis, assessing its feature transferability to hyperspectral imagery without additional training.

\textbf{Hyperspectral Images (HSI) Land-Cover Mapping.}
Hyperspectral imagery (HSI) enables fine-grained land-cover classification by capturing rich spectral information across hundreds of contiguous bands \cite{prasad2014segmented, wu2016dirichlet, li2012locality}. Recent methods adopt multimodal fusion (e.g., HSI with RGB or SAR) \cite{wang2022hyperspectral} and transformer-based architectures \cite{hong2021spectralformer}, but rely heavily on full supervision, where performance depends on the availability of labeled data. In contrast, we are the first to leverage a universal pre-trained diffusion model for HSI land-cover mapping, extracting strong transferable features without task-specific training and significantly reducing label requirements.

\section{Proposed Methodology}

\begin{figure*}[t]
\centering
\includegraphics[width=\textwidth]{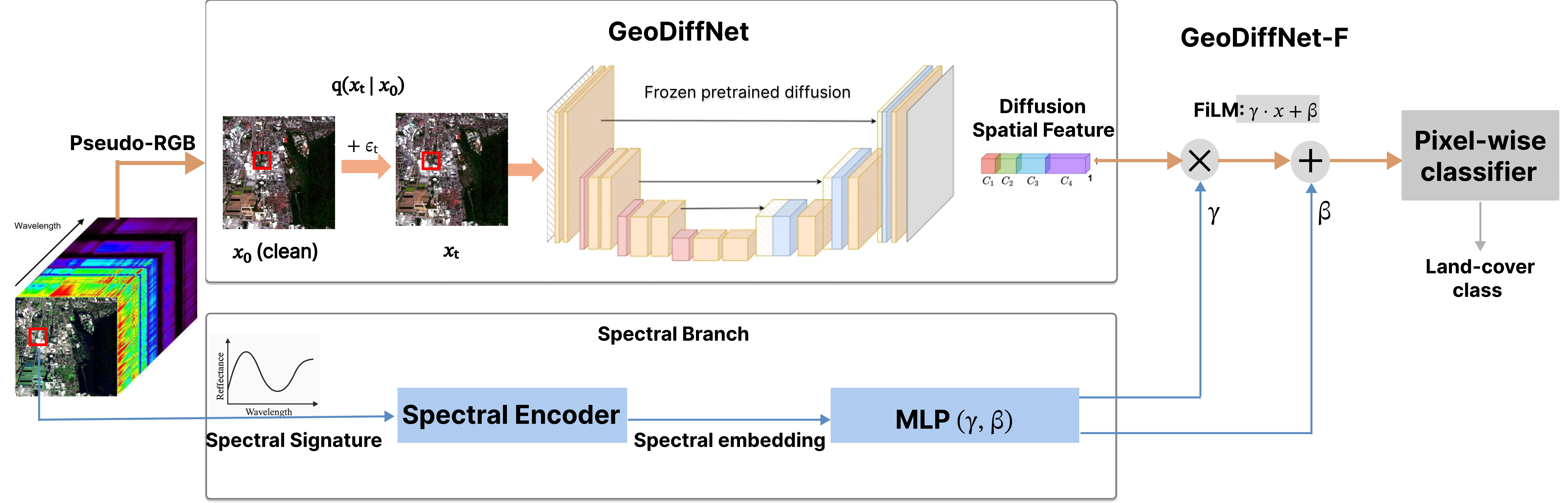}
\vskip -0.1in

    \caption{
        \textbf{Workflow of GeoDiffNet and GeoDiffNet-F.} 
        GeoDiffNet extracts low-level spatial features from RGB-like patches using a frozen pretrained diffusion model. A lightweight MLP is applied to each pixel for classification. GeoDiffNet-F further incorporates spectral context by encoding per-pixel reflectance signals into spectral embeddings, which are used to regress scaling (\( \gamma \)) and shifting (\( \beta \)) vectors through an MLP. These vectors condition the spatial features via a FiLM layer, enabling adaptive cross-modal fusion for land-cover classification.
    }

\label{fig:workflow}
\vskip -0.1in
\end{figure*}

\subsection{Overview of GeoDiffNet and GeoDiffNet-F}
Our framework consists of two complementary branches, as illustrated in \cref{fig:workflow}. The \textbf{GeoDiffNet} branch focuses on spatial feature extraction by leveraging a frozen diffusion model pretrained on natural RGB images. Specifically, we extract per-pixel features from low-level decoder layers at low denoising timesteps (e.g., T=0, 50, 100), which are shown to capture meaningful local structure even under resolution constraints. Each hyperspectral image is divided into overlapping pseudo-RGB patches (e.g., 64×64 with stride 32), and the extracted features are passed through a lightweight MLP for adaptation.

In parallel, the \textbf{spectral branch} encodes each pixel's full spectral signature using a dedicated spectral encoder, followed by an MLP that predicts FiLM parameters (scaling $\gamma$ and shifting $\beta$). These modulation parameters are used to condition the spatial features through a FiLM layer, enabling dynamic feature adaptation across modalities.

Finally, the \textbf{GeoDiffNet-F} module performs adaptive multimodal fusion. The modulated spatial features are passed to a 2-layer MLP for final pixel-wise land-cover classification. This design allows the model to benefit from both local spatial cues captured by the diffusion model and detailed spectral information unique to hyperspectral imagery.

\subsection{Diffusion Model for Spatial Feature Extraction}
To extract spatial features from hyperspectral images (HSI), we first select three spectral bands corresponding to the red, green, and blue wavelengths to construct a pseudo-RGB image. We then employ a pre-trained diffusion model (trained on ImageNet)~\cite{dhariwal2021diffusion,nichol2021glide,ranftl2021vision}. This model is based on a U-Net architecture, consisting of an encoder and a decoder. The decoder integrates information from the encoder through skip connections and contains 12 layers with varying resolutions and channel sizes, including attention mechanisms at specific scales to capture both local and global dependencies. We choose a $64 \times 64$ input resolution to align with the patch-based nature of land-cover mapping, where large images are divided into small patches for dense classification. This resolution not only reduces computational cost, but also matches the training scale of the diffusion model, which is optimized to reconstruct local spatial structures---making it particularly effective in low-texture or homogeneous regions common in HSI data. Additional implementation details about the pre-trained diffusion model are provided in \cref{sec:appendixD}.

A diffusion model involves two process: inversion (forward) and reversion (generation).
At \( t = 0 \), we extract the feature representation of the clean image \( x_0 \). Using the forward process, we can directly compute the noisy image \( x_t \) at timestep \( t \) as follows:
\begin{equation}
  x_t = \sqrt{\bar{\alpha}_t} x_0 + \sqrt{1 - \bar{\alpha}_t} \epsilon, \quad \epsilon \sim \mathcal{N}(0, 1),
  \label{eq:noisy_image}
\end{equation}
with the corresponding conditional distribution:
\begin{equation}
  q(x_t|x_0) := \mathcal{N}(x_t; \sqrt{\bar{\alpha}_t} x_0, (1 - \bar{\alpha}_t) I).
  \label{eq:conditional_distribution}
\end{equation}
Here, \( \bar{\alpha}_t \) represents the cumulative noise schedule up to timestep \( t \).

The forward process is used because the equation allows us to calculate \( x_t \) and its distribution \( q(x_t | x_0) \) from \( x_0 \) in a single step. This contrasts with the progressive generation process, which requires iterative computation of intermediate states, making the forward process more efficient for feature extraction in large-scale image processing.

Choosing the appropriate timestep \( t \) is crucial. Some studies suggest that early timesteps balance the original image and noise, providing richer feature representations \cite{baranchuk2021label}. However, the optimal timestep is still under exploration. While \( t = 0 \) may retain original features best \cite{xu2023open}, adding a bit of noise might enhance feature extraction \cite{luo2023diffusion}. We will first experiment with \( t = 0 \), 50, and 100 in \cref{subsec: low-level evaluation} and then conduct an ablation study to evaluate its efficacy on transferability of different timestep in \cref{ablation:transferbility}.

\subsection{Pre-trained Diffusion Model for Geospatial Imagery }

Given that geospatial images exhibit significant disparities from the natural images typically used to pre-train diffusion models, a key question arises: can these pre-trained diffusion models effectively extract the spatial features needed from geospatial imagery?

\textbf{Low-Level Features}: These are captured by the initial layers of the network and typically include basic patterns such as edges, textures, and simple shapes. In the context of the U-Net decoder, this would correspond to a layer close to the output (e.g. layer 9-11, with layer numbered from bottom to up).

\textbf{High-Level Features}: These are captured by the deeper layers of the network and involve more complex and abstract representations, such as parts of objects or entire objects. In the context of the U-Net decoder,this would generally correspond to decoder layers closer to the bottom of the UNet (e.g., layers 2-5)

According to \cite{long2015learning}, deep models' lower-level features have high transferability in domain adaptation. In the diffusion model U-Net architecture context, it's reasonable to conclude that the diffusion model's U-Net decoder, with its upper side (layer 9-11) corresponding to low-level features, can still effectively capture spatial features for geospatial image analysis tasks, despite the significant domain discrepancy.
\begin{figure}[t]
  \centering
  \vskip 0.1in

  \begin{minipage}[b]{0.9\linewidth}
    \centering
    \includegraphics[width=\linewidth]{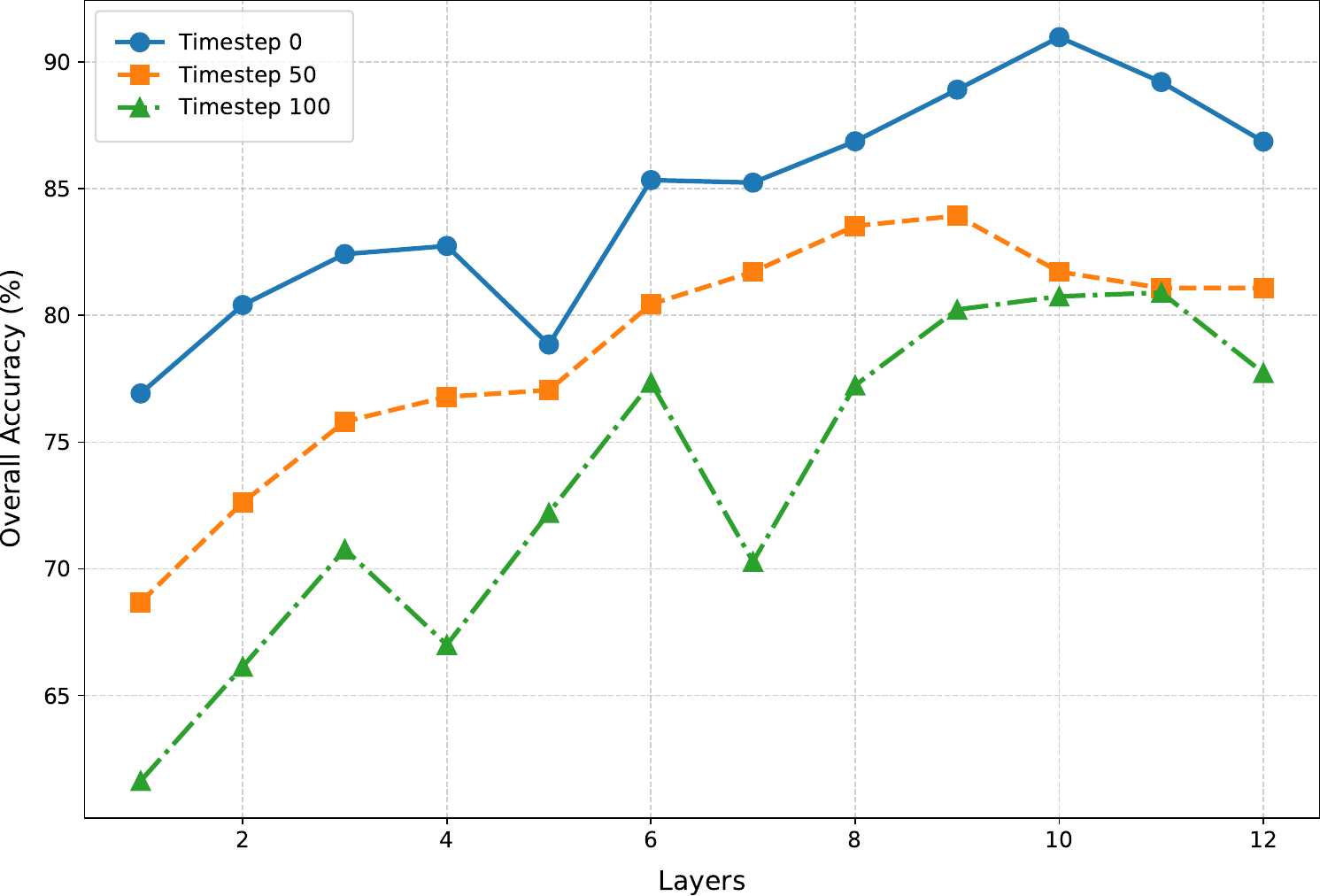}
    \vspace{0.3em}
    \small \textbf{(a)} Augsburg
  \end{minipage}

  \vspace{1em}

  \begin{minipage}[b]{0.9\linewidth}
    \centering
    \includegraphics[width=\linewidth]{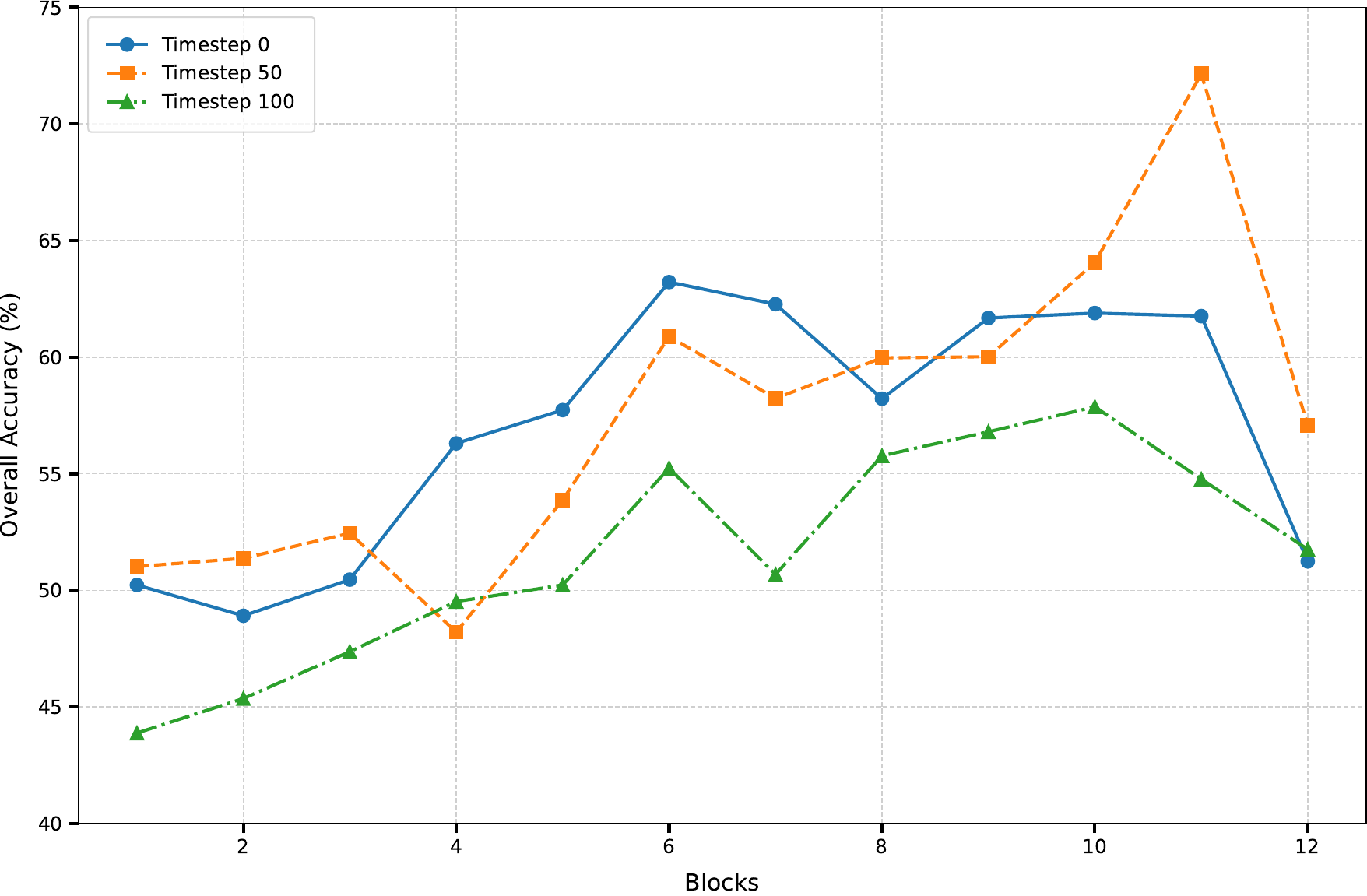}
    \vspace{0.3em}
    \small \textbf{(b)} Berlin
  \end{minipage}

  \caption{Both dataset performance metrics peak at \textbf{higher layers}, capturing \textbf{low-level features}. 
  (a) \textbf{Augsburg}: performance peaks at layer 10 (timestep 0). 
  (b) \textbf{Berlin}: performance peaks at layer 11 (timestep 50).}
  \label{fig:performance_over_layers}
  \vskip -0.15in
\end{figure}

\subsection{Efficacy of Low-Level Features}
Our goal is to evaluate the transferability of features extracted from a pre-trained diffusion model in the context of hyperspectral land-cover mapping. Specifically, we investigate whether spatial features obtained from shallow decoder layers and low-noise timesteps generalize better across domains.

Features extracted from a pre-trained diffusion model exhibit a dual hierarchy formed by both the model architecture and the denoising process. The spatial hierarchy is reflected across U-Net decoder layers: deeper layers encode high-level semantic features aligned with the pre-training domain, while shallower layers retain low-level spatial detail that is often more general and transferable. The temporal hierarchy arises across denoising timesteps: features at high-noise timesteps capture coarse, global structure, whereas those at low-noise timesteps recover finer, local details. Given the domain discrepancy between natural images used for pre-training and the target geospatial imagery, we hypothesize that low-level features (extracted from shallow decoder layers and low-noise timesteps) remain effective and transferable, as visualized in Appendix \ref{sec:Appendix_feature_visual}.

To examine spatial transferability, we extract pixel-level features from decoder layers 2 to 11, where lower layers (2–5) are known to encode high-level abstractions and upper layers (9–11) retain low-level spatial patterns. These features are obtained from pseudo-RGB input patches (\(64 \times 64\), stride 32), passed through the frozen diffusion model. A lightweight two-layer MLP is applied to classify each pixel based on the extracted feature vector.

This setup enables a systematic evaluation of feature transferability across both spatial and temporal axes of the diffusion model. We use overall accuracy (OA), average accuracy (AA), and kappa coefficient (KC) to assess which layer–timestep combinations yield the most transferable features. Full results and analysis are presented in \cref{ablation:transferbility}.

\subsection{Spectrally-Conditioned Spatial Modulation}

We propose a spectrally-conditioned spatial modulation mechanism using \textit{Feature-wise Linear Modulation (FiLM)} to adapt spatial features based on per-pixel spectral input. A frozen pretrained diffusion model is used to extract spatial feature vectors \( f_i^{\text{spatial}} \in \mathbb{R}^d \) from pseudo-RGB image patches. These spatial features capture structural and contextual patterns but lack detailed spectral information.

To incorporate spectral context, each pixel’s hyperspectral signature \( s_i \in \mathbb{R}^{b} \), where \( b \) denotes the number of spectral bands, is passed through a lightweight spectral encoder to produce a compact embedding. This embedding is then processed by a separate MLP to regress the FiLM modulation parameters: a scaling vector \( \gamma(s_i) \in \mathbb{R}^d \) and a shifting vector \( \beta(s_i) \in \mathbb{R}^d \). These parameters are applied to the corresponding spatial feature vector as follows:
\begin{equation}
\hat{f}_i = \gamma(s_i) \cdot f_i^{\text{spatial}} + \beta(s_i).
\label{eq:film_modulation}
\end{equation}
This pixel-wise conditioning enables the model to dynamically adapt spatial representations using the spectral characteristics of each pixel. Compared to traditional fusion methods such as concatenation or summation, FiLM allows for more flexible and learnable cross-modal interaction, leading to improved performance in land-cover classification tasks under limited supervision. An overview of this fusion strategy is illustrated in \cref{fig:workflow}.

\noindent \textbf{GeoDiffNet-F: Extending Diffusion Features with Spectral Reflectance Information.}
Building on the spectrally-conditioned spatial modulation framework described above, we define GeoDiffNet-F as our final architecture for pixel-wise land-cover classification. Each pixel’s hyperspectral signature is first encoded through a shallow MLP network, followed by an MLP that regresses FiLM modulation parameters. These parameters adapt the spatial features extracted from a frozen diffusion model, enabling spectral-to-spatial conditioning. The modulated features are passed to a lightweight classifier for prediction. As shown in \cref{fig:workflow}, this formulation enhances spatial representations using spectral context, resulting in improved performance under domain shift and low-label regimes.

\section{Experimental Setup and Results}
\begin{table}[h]
\caption{Training and testing samples for the \textbf{Augsburg} dataset.}
\label{tab:Augsburg_train_test_splits}
\begin{center}
\begin{small}
\begin{sc}
\begin{tabular}{lrr}
\toprule
Class & Train Count & Test Count \\
\midrule
Forest           & 146 & 13,361 \\
Residential Area & 264 & 30,065 \\
Industrial Area  & 21  & 3,830  \\
Low Plants       & 248 & 26,609 \\
Allotment        & 52  & 523    \\
Commercial Area  & 7   & 1,638  \\
Water            & 23  & 1,507  \\
\midrule
\textbf{Total}   & \textbf{761} & \textbf{77,533} \\
\bottomrule
\end{tabular}
\end{sc}
\end{small}
\end{center}
\vskip -0.1in
\end{table}

\begin{table}[ht]
\caption{Training and testing samples for the \textbf{Berlin} dataset.}
\label{tab:Berlin_train_test_splits}
\begin{center}
\begin{small}
\begin{sc}
\begin{tabular}{lrr}
\toprule
Class & Train Count & Test Count \\
\midrule
Forest            & 443 & 54,511 \\
Residential Area  & 423 & 268,219 \\
Industrial Area   & 499 & 19,067 \\
Low Plants        & 376 & 58,906 \\
Soil              & 331 & 17,095 \\
Allotment         & 280 & 13,025 \\
Commercial Area   & 298 & 24,526 \\
Water             & 170 & 6,502 \\
\midrule
\textbf{Total}    & \textbf{2,820} & \textbf{461,851} \\
\bottomrule
\end{tabular}
\end{sc}
\end{small}
\end{center}
\vskip -0.1in
\end{table}

\subsection{Dataset}
To validate the proposed method, we use two publicly available hyperspectral datasets,Augsburg and Berlin,capturing urban and rural regions in Germany \cite{hong2021multimodal}.

The Augsburg dataset was collected using the HySpex sensor and contains 180 spectral bands covering wavelengths from 0.4–2.5 µm. It has a spatial resolution of 30\,m GSD and an image size of $332 \times 485$ pixels.

The Berlin dataset, synthesized from HyMap HSI data to resemble EnMAP spectral characteristics, contains 244 bands over the same spectral range, with a resolution of 30\,m GSD and dimensions of $797 \times 220$ pixels.

We adopt the original train/test splits from \cite{hong2021multimodal}, summarized in \cref{tab:Augsburg_train_test_splits} and \cref{tab:Berlin_train_test_splits}, to ensure consistency and enable direct comparison with prior work \cite{wang2022hyperspectral}.


\subsection{Implementation Details}
\textbf{Pre-trained Diffusion Model.} We use the pre-trained diffusion model from \cite{dhariwal2021diffusion} with a $64 \times 64$ patch size. This choice leverages the pretrained backbone without introducing additional trainable parameters, while providing abundant spatial context—over $30\times$ larger than the typical $11 \times 11$ HSI patches used in geospatial tasks~\cite{ahmad2020fast}—allowing the model to capture long-range dependencies more effectively.

To utilize a pre-trained diffusion model, two decisions must be made: selecting between the encoder or decoder, and choosing between the forward or reverse process. We opt for the decoder, as in the U-Net architecture, the decoder integrates feature maps from the encoder via skip connections \cite{baranchuk2021label}.
\textbf{Forward process} is opted for feature extraction because it operates in a single timestep, making it more efficient than the progressive reverse process while achieving comparable performance \cite{zhong2024diffusion,luo2023diffusion}.
\begin{figure*}[t]
  \centering
  \begin{minipage}{.19\textwidth}
    \includegraphics[height=0.53\textheight]{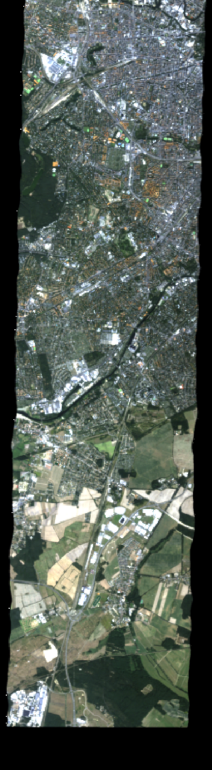}
    \centering \small (a) RGB
  \end{minipage}
  \hspace{0.1em}
  \begin{minipage}{.19\textwidth}
    \includegraphics[height=0.53\textheight]{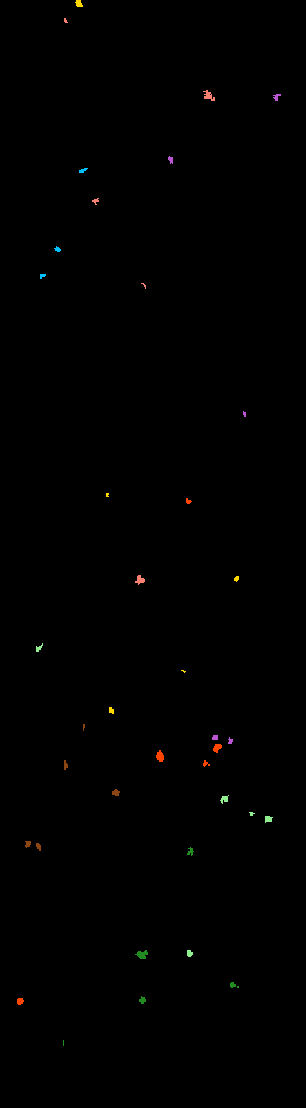}
    \centering \small (b) Train label
  \end{minipage}
  \hspace{0.1em}
  \begin{minipage}{.19\textwidth}
    \includegraphics[height=0.53\textheight]{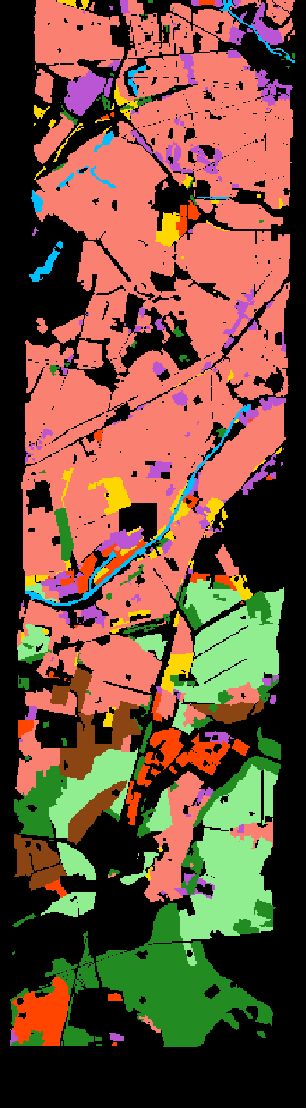}
    \centering \small (c) Ground-Truth
  \end{minipage}
  \hspace{0.1em}
  \begin{minipage}{.19\textwidth}
    \includegraphics[height=0.53\textheight]{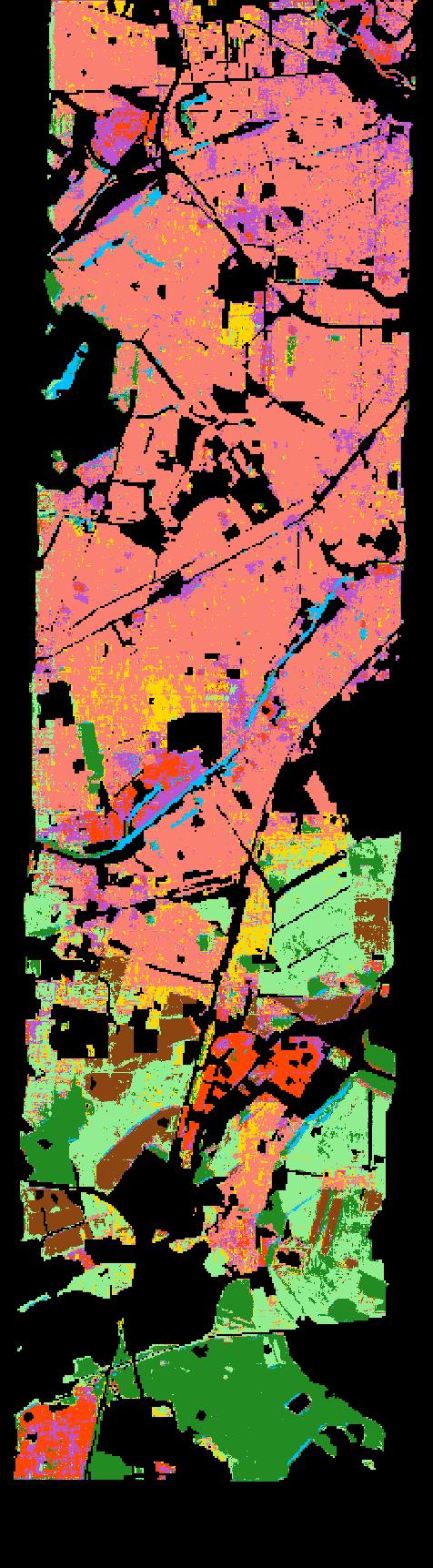}
    \centering \small (d) \textbf{GeoDiffNet}
  \end{minipage}
  \hspace{0.1em}
  \begin{minipage}{.19\textwidth}
    \includegraphics[height=0.53\textheight]{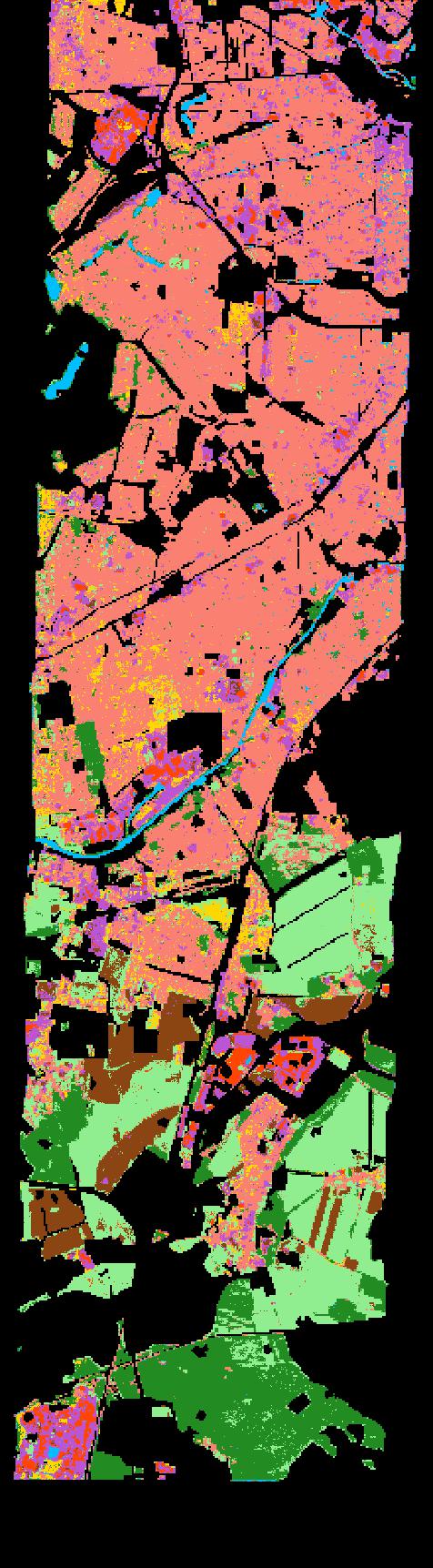}
    \centering \small (e) \textbf{GeoDiffNet-F}
  \end{minipage}

  \vspace{0.53em}
  \includegraphics[width=\textwidth]{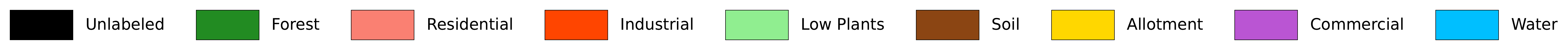}

  \caption{Visualization on \textbf{Berlin} HSI: (a) RGB image, (b) Training label map, (c) Ground-Truth (test label), (d) \textbf{GeoDiffNet} output, and (e) \textbf{GeoDiffNet-F}.}
  \label{fig:berlin_spatial_sota}
\end{figure*}

\begin{table*}[t]
\caption{Performance comparison on the \textbf{Berlin} dataset.
TBCNN~\cite{xu2018multisource}, S2FL~\cite{hong2021multimodal},
ContextCNN~\cite{lee2017going}, DFINet~\cite{gao2021hyperspectral},
and MIFNet~\cite{wang2022hyperspectral} are prior methods.
\textbf{GeoDiffNet} uses spatial features from diffusion layer 11 of pseudo-RGB HSI;
\textbf{GeoDiffNet-F} incorporates fused HSI spectral and spatial features.}
\label{tab:Berlin_icml}
\centering
\scriptsize
\setlength{\tabcolsep}{2pt}
\resizebox{\textwidth}{!}{%
\begin{tabular}{@{}lcccccccccccc@{}}
\toprule
\textbf{Method} & \textbf{Modality} & \textbf{Forest} & \textbf{Res.} &
\textbf{Indust.} & \textbf{L.Plants} & \textbf{Soil} & \textbf{Allot.} &
\textbf{Comm.} & \textbf{Water} & \textbf{OA (\%)} & \textbf{AA (\%)} & \textbf{KC (\%)} \\
\midrule
TBCNN             & HSI         & 71.52 & 60.80 & 69.58 & 68.57 & 80.39 & 97.55 & 35.25 & 82.77 & 63.85 & 69.55 & 39.94 \\
S2FL              & HSI+SAR     & 83.30 & 57.39 & 48.53 & 77.16 & 83.84 & 57.05 & 31.02 & 61.57 & 62.23 & 62.48 & 48.77 \\
ContextCNN        & HSI+SAR     & 77.22 & 63.69 & 61.44 & 73.77 & 87.22 & 82.88 & 31.13 & 74.24 & 66.31 & 68.95 & 54.03 \\
DFINet            & HSI+SAR     & 68.95 & 67.52 & 43.42 & 81.77 & 75.58 & 80.05 & 40.94 & 79.87 & 67.93 & 67.26 & 55.22 \\
MIFNet            & HSI+SAR     & 68.77 & 76.90 & 50.75 & 81.10 & 65.59 & 75.69 & 29.96 & 82.96 & 72.54 & 66.47 & 59.81 \\
\textbf{GeoDiffNet}   & HSI     & 75.35 & 77.82 & 53.41 & 71.50 & 75.12 & 36.86 & 40.41 & 54.74 & 72.15 & 60.65 & 58.50 \\
\textbf{GeoDiffNet-F} & HSI & 80.21 & 79.33 & 26.01 & 84.80 & 82.15 & 28.68 & 41.53 & 67.99 & \textbf{74.44} & 61.34 & \textbf{61.32} \\
\bottomrule
\end{tabular}
}
\end{table*}

\textbf{Data Preparation.}
To prepare pseudo-RGB inputs for \textsc{GeoDiffNet}, we selected three representative spectral bands ---- bands 40, 30, and 15 for Berlin, and bands 21, 11, and 6 for Augsburg ----since they approximately correspond to red, green, and blue wavelengths in the visible spectrum. This choice facilitates intuitive visualization and aligns with the RGB distribution seen during diffusion model pretraining.
 Hyperspectral images were divided into overlapping $64 \times 64$ patches with a stride of 32. Padding was applied to preserve spatial coverage and minimize edge artifacts.

\textbf{GeoDiffNet.}
Each $64 \times 64$ pseudo-RGB patch was processed using a frozen pretrained diffusion model. Decoder activations from layers 2 to 11 were resized to patch resolution for pixel-level alignment. For each labeled pixel, the corresponding spatial feature was used to train a two-layer MLP classifier, enabling evaluation across different layers and timesteps.

\textbf{Spectral Branch.}
Each labeled pixel’s spectral reflectance vector (180 bands for Augsburg, 244 for Berlin) was passed through a shallow MLP encoder. The output was used to regress FiLM parameters—scaling ($\gamma$) and shifting ($\beta$)—for feature modulation.

\textbf{GeoDiffNet-F.}
We applied FiLM modulation to the frozen spatial features using the spectral FiLM parameters. The modulated features were passed through a two-layer MLP for final pixel-wise classification. Only the spectral branch and classifier were trained; the diffusion model remained frozen.

\textbf{Training and Inference }
GeoDiffNet was trained to evaluate the effectiveness of frozen diffusion features, while GeoDiffNet-F was trained by optimizing the spectral branch---including the encoder and FiLM parameter regressor—and the classification layers, while keeping the diffusion-based spatial backbone frozen. We used a learning rate of 0.003, batch size of 64, and trained for up to 10 epochs with early stopping if no validation improvement occurred within 1000 iterations.

During inference, large geospatial images were divided into 64 × 64 patches with a stride of 32, creating overlapping regions. Pixel-wise predictions in overlapping areas were aggregated using max-voting to ensure smooth and accurate classification.

\begin{table*}[t]
\caption{Performance comparison on the \textbf{Augsburg} dataset.
TBCNN~\cite{xu2018multisource}, S2FL~\cite{hong2021multimodal},
ContextCNN~\cite{lee2017going}, DFINet~\cite{gao2021hyperspectral},
and MIFNet~\cite{wang2022hyperspectral} are prior methods.
\textbf{GeoDiffNet} uses spatial features from diffusion layer 10 of pseudo-RGB HSI; 
\textbf{GeoDiffNet-F} incorporates fused HSI spectral and spatial features.}
\label{tab:augsburg_icml}
\centering
\scriptsize
\setlength{\tabcolsep}{2pt}
\resizebox{\textwidth}{!}{%
\begin{tabular}{@{}lccccccccccc@{}}
\toprule
\textbf{Method} & \textbf{Modality} & \textbf{Forest} & \textbf{Res.} &
\textbf{Indust.} & \textbf{L.Plants} & \textbf{Allot.} & \textbf{Comm.} &
\textbf{Water} & \textbf{OA (\%)} & \textbf{AA (\%)} & \textbf{KC (\%)} \\
\midrule
TBCNN             & HSI         & 94.71 & 96.37 & 69.30 & 81.58 & 62.52 & 12.70 & 16.39 & 86.12 & 61.94 & 80.24 \\
S2FL              & HSI+SAR     & 88.80 & 86.36 & 38.90 & 90.53 & 68.64 & 8.97  & 47.45 & 83.36 & 61.38 & 76.26 \\
ContextCNN        & HSI+SAR     & 94.57 & 97.25 & 51.46 & 86.25 & 56.02 & 13.68 & 21.57 & 87.24 & 60.11 & 81.82 \\
DFINet            & HSI+SAR     & 95.38 & 95.84 & 69.79 & 86.65 & 64.05 & 13.86 & 28.47 & 88.06 & 64.86 & 82.98 \\
MIFNet            & HSI+SAR     & 92.28 & 96.53 & 59.53 & 90.79 & 59.46 & 17.58 & 51.43 & 89.21 & 66.80 & 84.53 \\
\textbf{GeoDiffNet}   & HSI     & 92.78 & 98.04 & 61.46 & 95.98 & 86.42 & 6.35  & 14.33 & 90.98 & 65.05 & 86.82 \\
\textbf{GeoDiffNet-F} & HSI     & 91.77 & 98.02 & 64.18 & 95.67 & 86.42 & 12.33 & 28.80 & \textbf{91.23} & \textbf{68.17} & \textbf{87.26} \\
\bottomrule
\end{tabular}
}
\end{table*}

\begin{figure*}[t]
  \centering

  \begin{minipage}{0.19\textwidth}
    \rotatebox{90}{\includegraphics[width=0.21\textheight]{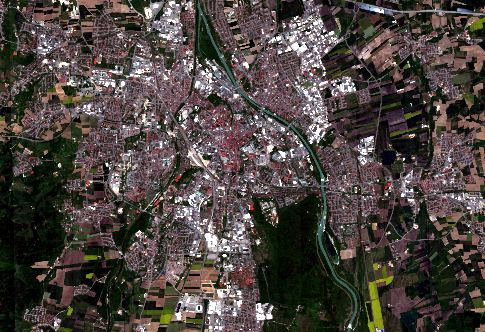}}
    \centering \small(a) Pseudo-RGB
  \end{minipage}
  \hspace{0.05em}
  \begin{minipage}{0.19\textwidth}
    \rotatebox{90}{\includegraphics[width=0.21\textheight]{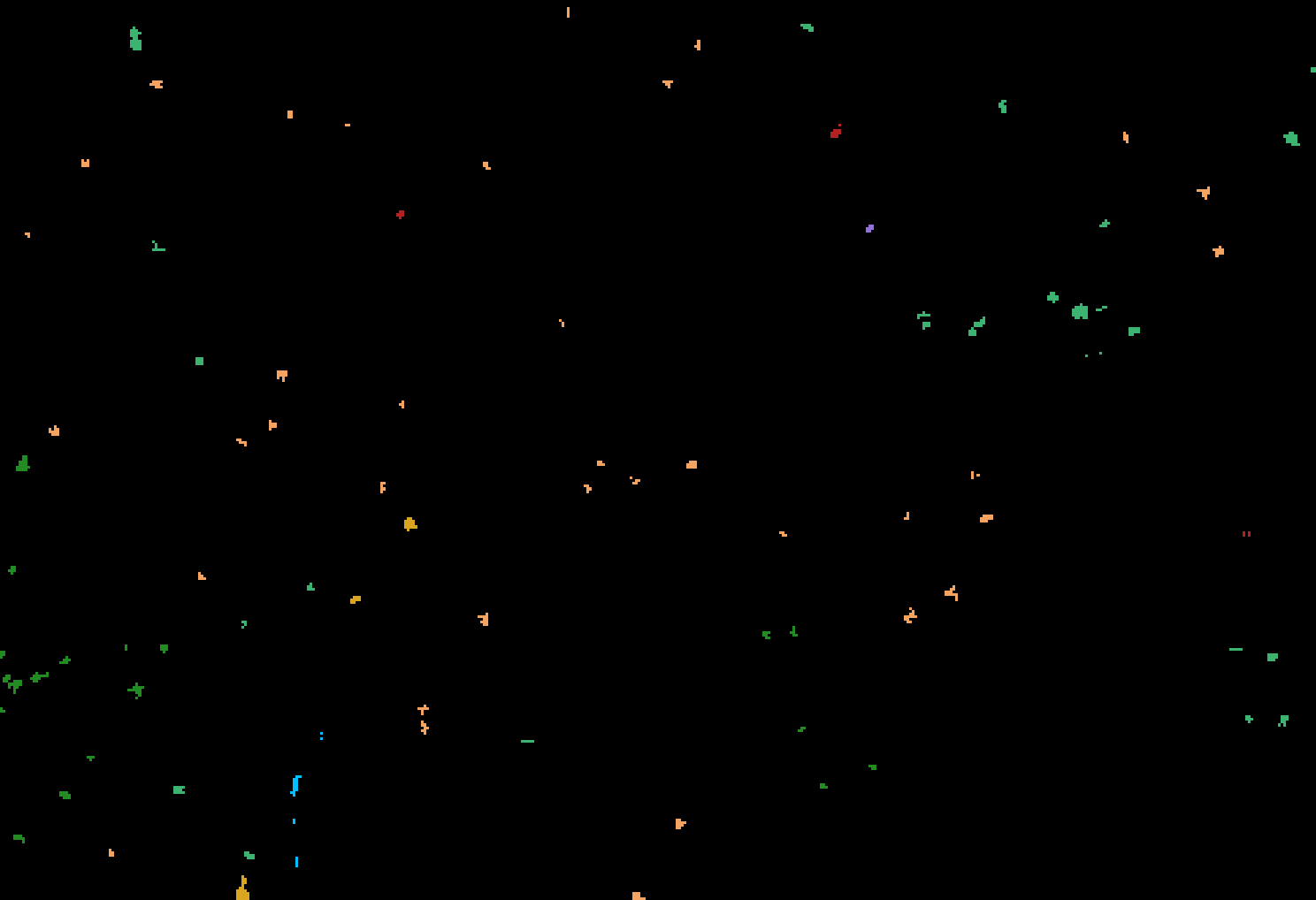}}
    \centering \small(b) Train label
  \end{minipage}
  \hspace{0.05em}
  \begin{minipage}{0.19\textwidth}
    \rotatebox{90}{\includegraphics[width=0.21\textheight]{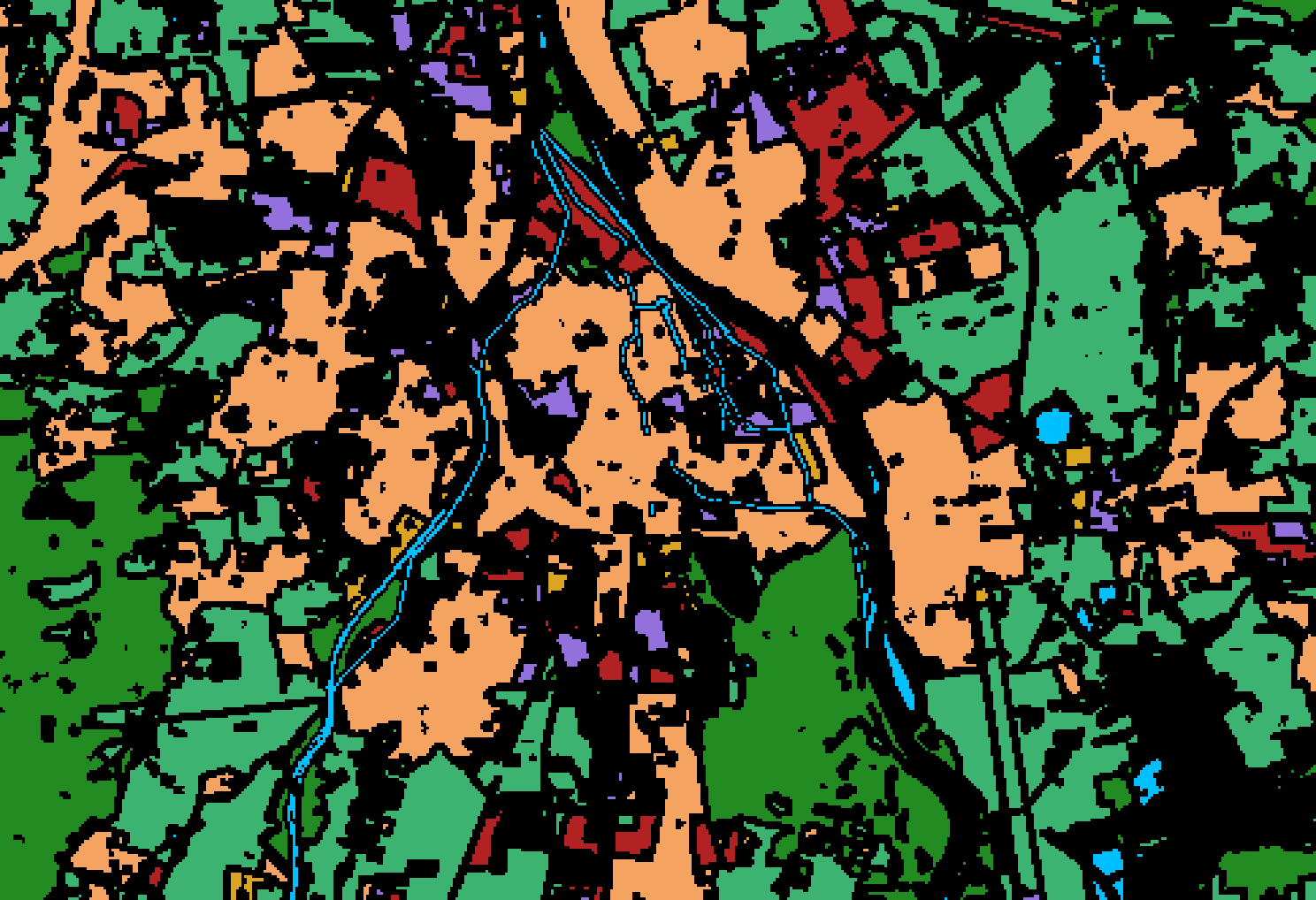}}
    \centering \small(c) Ground-Truth
  \end{minipage}
  \hspace{0.05em}
  \begin{minipage}{0.19\textwidth}
    \rotatebox{90}{\includegraphics[width=0.21\textheight]{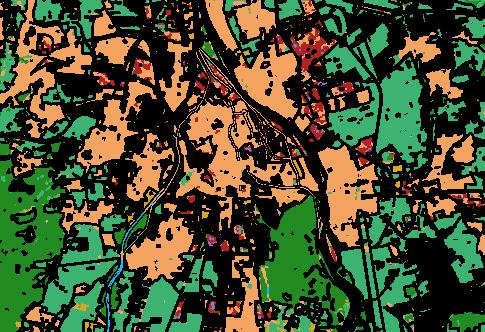}}
    \centering \small\textbf{(d) GeoDiffNet}
  \end{minipage}
  \hspace{0.05em}
  \begin{minipage}{0.19\textwidth}
    \rotatebox{90}{\includegraphics[width=0.21\textheight]{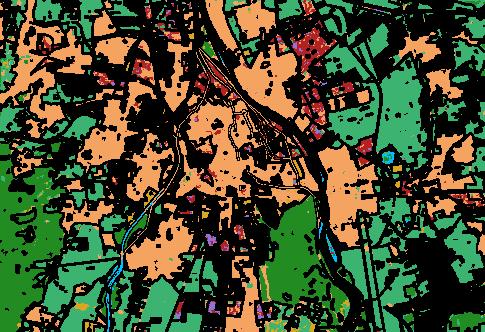}}
    \centering \small\textbf{(e) GeoDiffNet-F}
  \end{minipage}

  \vspace{0.5em}
  \includegraphics[width=\textwidth]{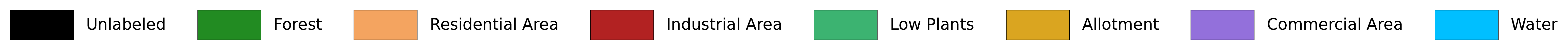}

  \caption{Visualization on \textbf{Augsburg} HSI: (a) Pseudo-RGB image, (b) Training label map, (c) Ground-truth, (d) \textbf{GeoDiffNet}, and (e) \textbf{GeoDiffNet-F}.}
  \label{visualization_Augsburg_spatial_sota}
\end{figure*}

\subsection{Evaluating the Efficacy of Low-Level Features}
\label{subsec: low-level evaluation}

\textbf{Qualitatively}, as demonstrated in \cref{fig:berlin_spatial_sota} and \cref{visualization_Augsburg_spatial_sota}, using low-level feature layers, such as layer 11 for Berlin and layer 10 for Augsburg, GeoDiffNet's segmentation results are highly effective.

Compared to the ground truth maps, using pre-trained diffusion model-extracted low-level features only, GeoDiffNet produces sharp and well-defined boundaries between different land cover classes, accurately capturing intricate details. This highlights GeoDiffNet's capability to enhance spatial resolution and classification precision. More visualizations across different layers of comparison can be found in \cref{sec:appendixA}.

\textbf{Quantitively}, from \cref{tab:augsburg_icml} and \cref{tab:Berlin_icml} , GeoDiffNet's low-level features, extracted using only HSI RGB 3 bands at higher layers, outperform several SOTA models that even rely on additional modalities in conjunction with HSI. For the Augsburg dataset, GeoDiffNet achieves an overall accuracy (OA) of 90.98\% and an average accuracy (AA) of 65.05\%, surpassing MIFNet and DFINet. For the Berlin dataset, GeoDiffNet records an OA of 72.15\% and an AA of 60.74\%, exceeding ContextCNN and comparable to DFINet.

Despite the significant disparity presented by geospatial imagery compared to the images that the pre-trained model has seen, as shown in  GeoDiffNet's low-level features are highly effective. This proves our hypothesis that extracted low-level features have great transferability.

These results underscore the superior performance of GeoDiffNet's higher-layer low-level features, highlighting the model's ability to achieve high classification accuracy and distinguish between similar land cover classes only using a subset of available channels in HSI, in contrast to other models that utilize the full spectrum and some that even require multiple modalities in addition to HSI.
\subsection{Efficacy of GeoDiffNet-F}

\textbf{Qualitatively}, as shown in 
\cref{visualization_Augsburg_spatial_sota,fig:berlin_spatial_sota}, 
GeoDiffNet-F exhibits clear improvements over GeoDiffNet for both 
Augsburg and Berlin datasets.

\textbf{Quantitatively}, GeoDiffNet-F with FiLM-based spectral fusion 
achieves the highest overall accuracy (OA) and Kappa coefficient (KC) 
on both datasets, as shown in \cref{tab:augsburg_icml,tab:Berlin_icml}. 
Although its AA on Berlin is slightly lower than MIFNet, OA and KC 
provide a more balanced evaluation under class imbalance, highlighting 
the robustness of our method.

The strong performance of the GeoDiffNet-F fusion model is largely attributable to the spatial features extracted by GeoDiffNet. By effectively exploiting the spatial features inherent in HSI data, GeoDiffNet-F not only surpasses other SOTA fusion methods that depend on multiple data sources but also demonstrates the significant potential of GeoDiffNet-extracted spatial features in advancing hyperspectral image analysis. 

\subsection{Ablation Study: Timestep and Decoder Layer}
\label{ablation:transferbility}
To investigate critical design choices, we conducted a detailed ablation study on GeoDiffNet, focusing specifically on how timestep selection (noise levels) and decoder layer depth (spatial resolution) affect the quality and transferability of extracted spatial features.
\vspace{-9pt}

\paragraph{Impact of Timestep (Noise Level)} 
Diffusion models introduce varying noise levels at different timesteps during the forward process. Lower timesteps correspond to cleaner images, whereas higher timesteps introduce progressively more noise. Our analysis as illustrated clearly in \cref{sec:appendixB} indicates that early timesteps generally exhibit higher feature transferability due to proximity to the original data distribution, consistent with the \textit{Chain of Forgetting theorem} \cite{zhong2024diffusion}.

However, our empirical findings highlight a nuanced observation: while clean images at timestep~0 yield optimal performance for Augsburg, a small amount of added noise at timestep~50 enhances feature extraction for Berlin. This aligns with previous studies~\cite{xu2023open,luo2023diffusion}, which suggest that minimal noise can help retain critical spatial details, effectively balancing high-frequency information and smoothness for improved classification accuracy.
Thus, while early timestep,low noise has more transferability, optimal timestep selection is not universally minimal and depends heavily on dataset-specific characteristics, reinforcing the necessity of carefully tuning this parameter.

\noindent \textbf{Impact of Decoder Layer (Spatial Resolution) } 
As diffusion models decode from lower to higher resolutions through progressive layers, deeper decoder layers yield more spatially detailed and accurate features. Our experiments demonstrate that higher layers consistently provide more informative representations, resulting in improved classification accuracy and better generalization to geospatial imagery. Specifically, optimal spatial feature extraction was achieved at \textbf{layer 10 for Augsburg} and \textbf{layer 11 for Berlin} (see detailed quantitative and qualitative analyses in \cref{sec:appendixA}).

\section{Conclusion}
We demonstrate that pretrained diffusion models can effectively transfer spatial representations to hyperspectral imagery without domain-specific finetuning. GeoDiffNet achieves strong pixel-level classification performance using only lightweight classifiers and minimal labeled data. By introducing FiLM-based spectral modulation, GeoDiffNet-F further improves performance through dynamic spatial--spectral fusion. Our analysis highlights that early decoder layers and lower noise timesteps yield the most transferable features, underscoring the potential of diffusion features as robust, label-efficient representations for remote sensing.

\section*{Code Availability}
Our code is publicly available at \url{https://github.com/hutuhehe/diffusion_hyperspectral}.

\section*{Impact Statement}
This work investigates how pretrained diffusion models can be repurposed for hyperspectral land cover mapping with minimal labeled data. By reducing reliance on large domain-specific datasets, our method promotes more accessible and label-efficient solutions for remote sensing. These capabilities have potential applications in environmental monitoring, agriculture, and disaster response—especially in regions with limited annotation resources. We believe our findings encourage broader exploration of generative models for geospatial analysis, while presenting minimal foreseeable ethical or societal risks.


{\small
\bibliographystyle{IEEEtran}
\bibliography{mybib}
}
\newpage
\appendix
\onecolumn
\section*{\Large{Appendix}}
\addcontentsline{toc}{section}{Appendix}
\noindent In \cref{sec:Appendix_feature_visual}, we provide visualizations of the spatial features extracted by the diffusion model.

\noindent In \cref{sec:appendixA}, We provide a detailed analysis of model performance over different layers of the diffusion model.

\noindent In \cref{sec:appendixB}, we provide an analysis of model performance as a function of different timesteps in the diffusion model.

\noindent In \cref{sec:appendixD}, we summarize the pre-trained model used for feature extraction, detailing the resolution, channels, and attention at each decoder layer.

\noindent \textbf{Code Availability.} Our code is publicly available at\\ \url{https://github.com/hutuhehe/diffusion_hyperspectral}.

\renewcommand{\thesection}{\Alph{section}} 
\renewcommand{\thesubsection}{\thesection.\arabic{subsection}} 


\section{Diffusion Spatial Feature Visualization}
\label{sec:Appendix_feature_visual}

We present $k$-means clustering results ($k=6$) on decoder features extracted from Layers 6–11 of a pretrained diffusion model, across timesteps from $T=0$ to $T=200$. Clustering is performed on a $64{\times}64$ pseudo-RGB patch sampled from the Berlin hyperspectral dataset, centered on the Messe Berlin convention center. Due to the low spatial resolution and limited texture in the input, object boundaries are not clearly visible. For reference, we include a high-resolution satellite image from Google Earth (circa 2009) to provide context on the actual scene layout.

\begin{figure*}[h]
\centering
\setlength{\tabcolsep}{2pt}
\renewcommand{\arraystretch}{1.1}

\begin{tabular}{c|c|c|c|c|c}
\textbf{Reference} & \textbf{Layer} & \textbf{T=0} & \textbf{T=50} & \textbf{T=100} & \textbf{T=200} \\
\midrule

\multirow{3}{*}{
    \begin{tabular}{@{}c@{}}
    \includegraphics[height=1.1cm]{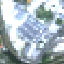} \\
    \scriptsize RGB patch\\ \scriptsize (64×64)
    \end{tabular}
  }

& \raisebox{0.4cm}{Layer 6} &

\includegraphics[width=0.14\linewidth]{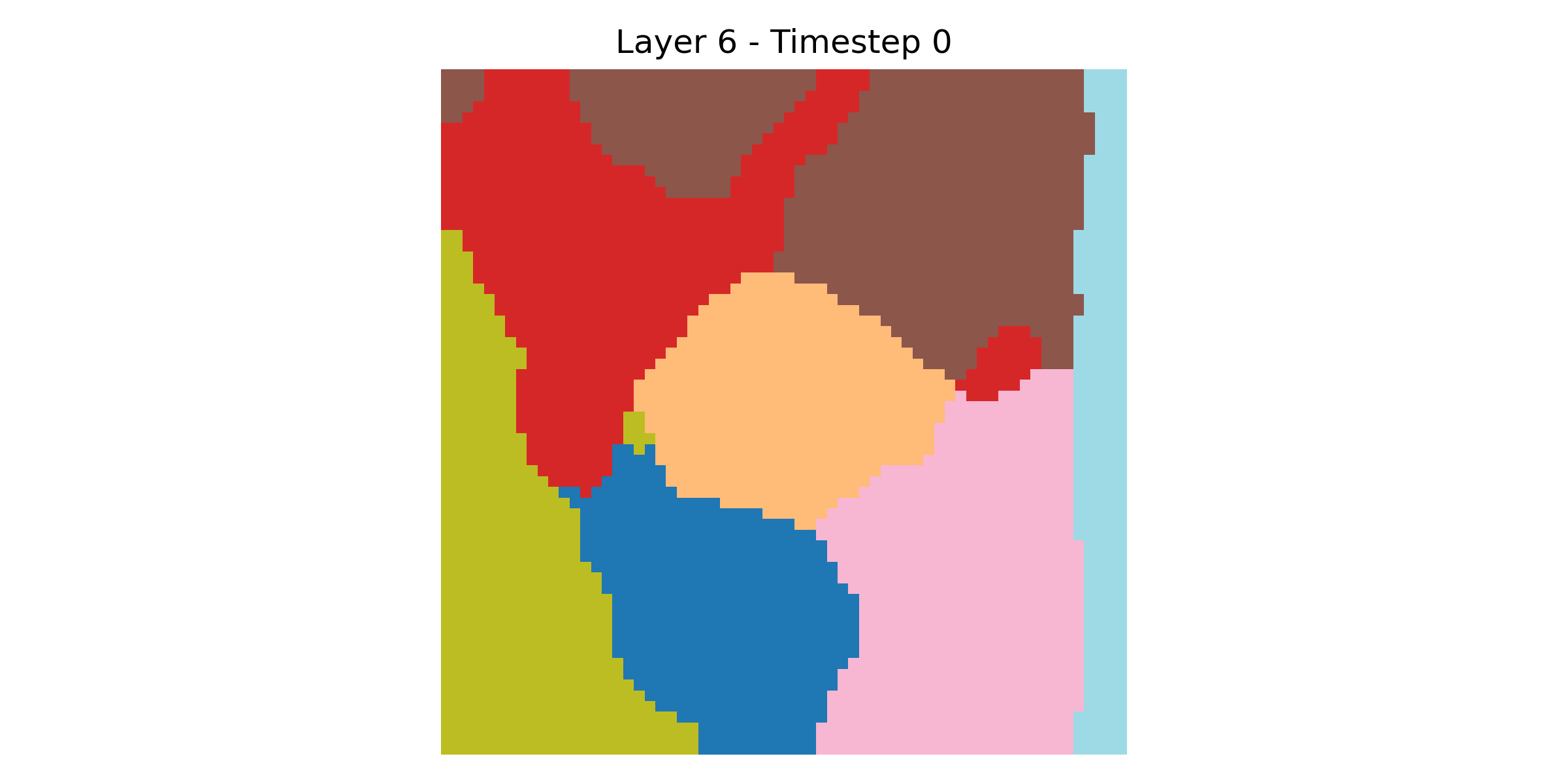} &
\includegraphics[width=0.14\linewidth]{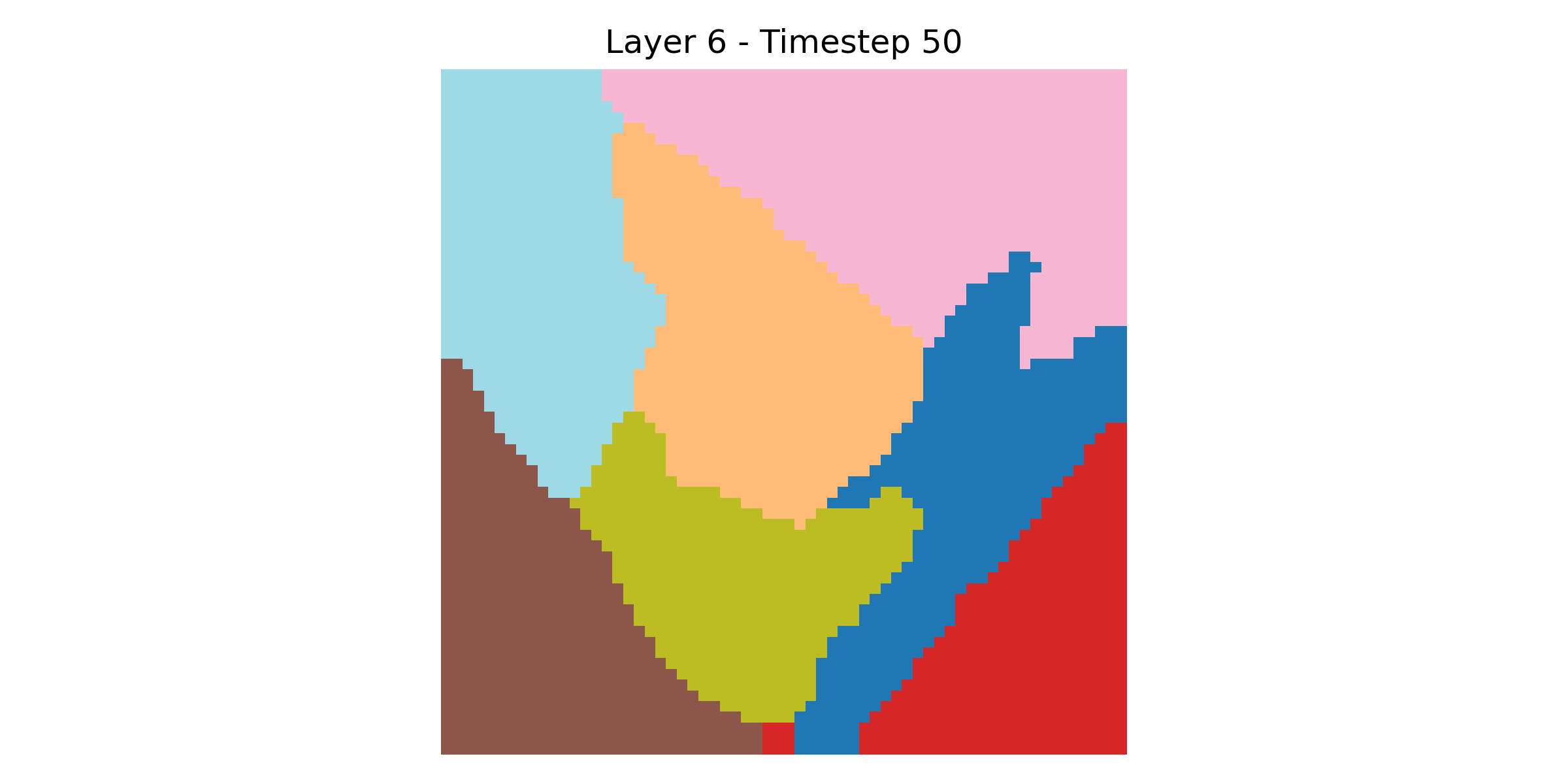} &
\includegraphics[width=0.14\linewidth]{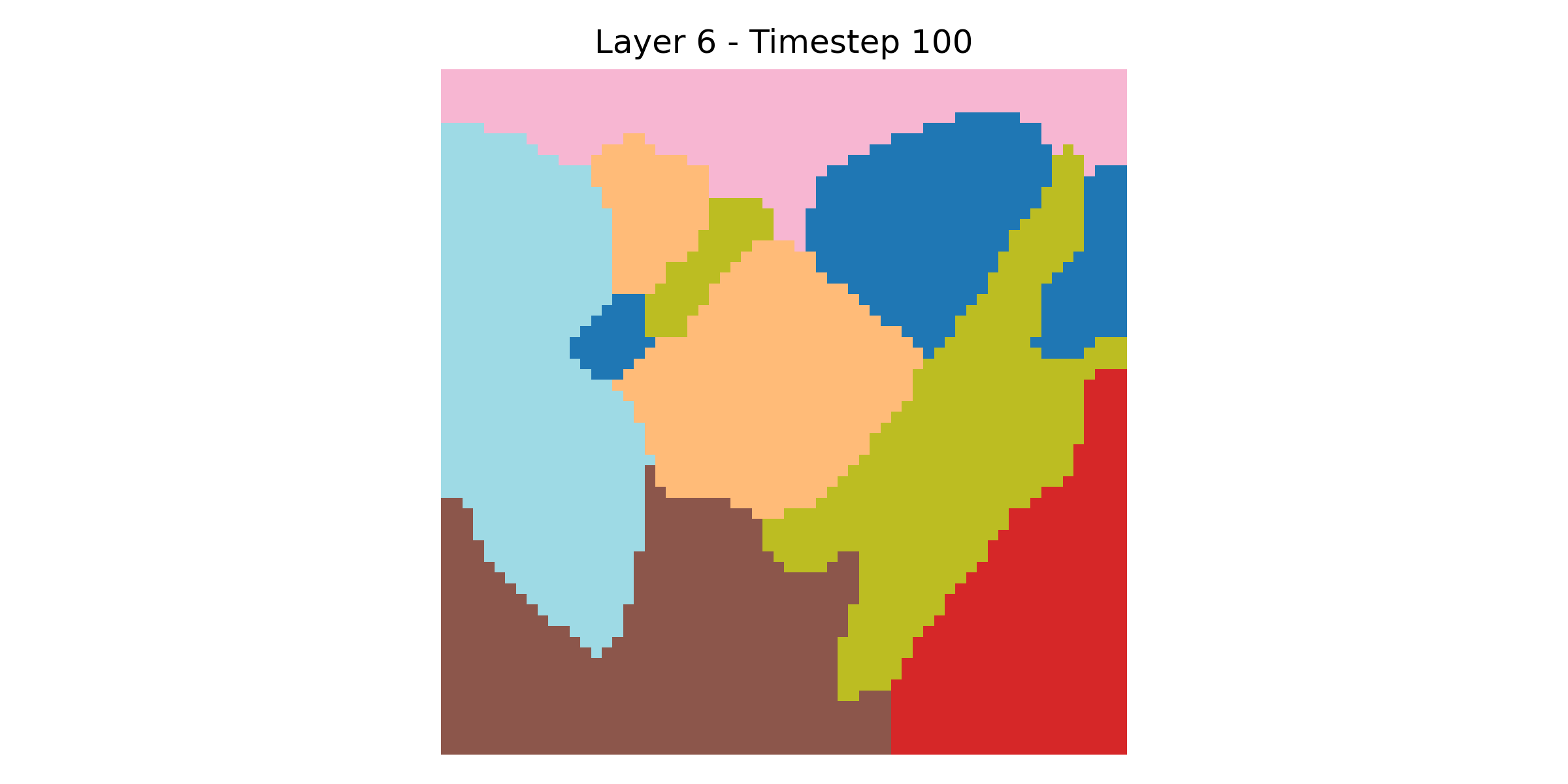} &
\includegraphics[width=0.14\linewidth]{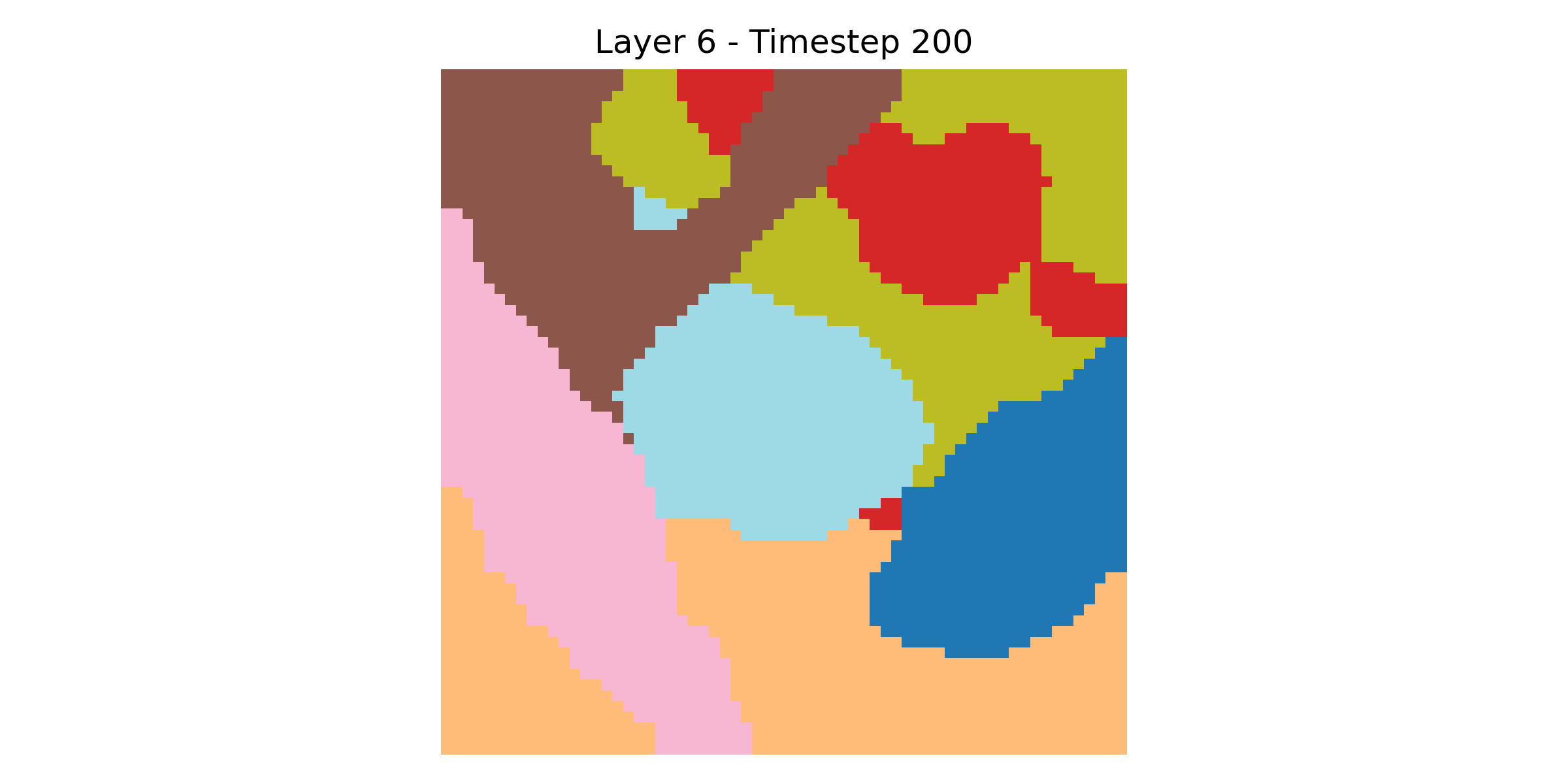} \\

&\raisebox{0.4cm}{Layer 7}  &
\includegraphics[width=0.14\linewidth]{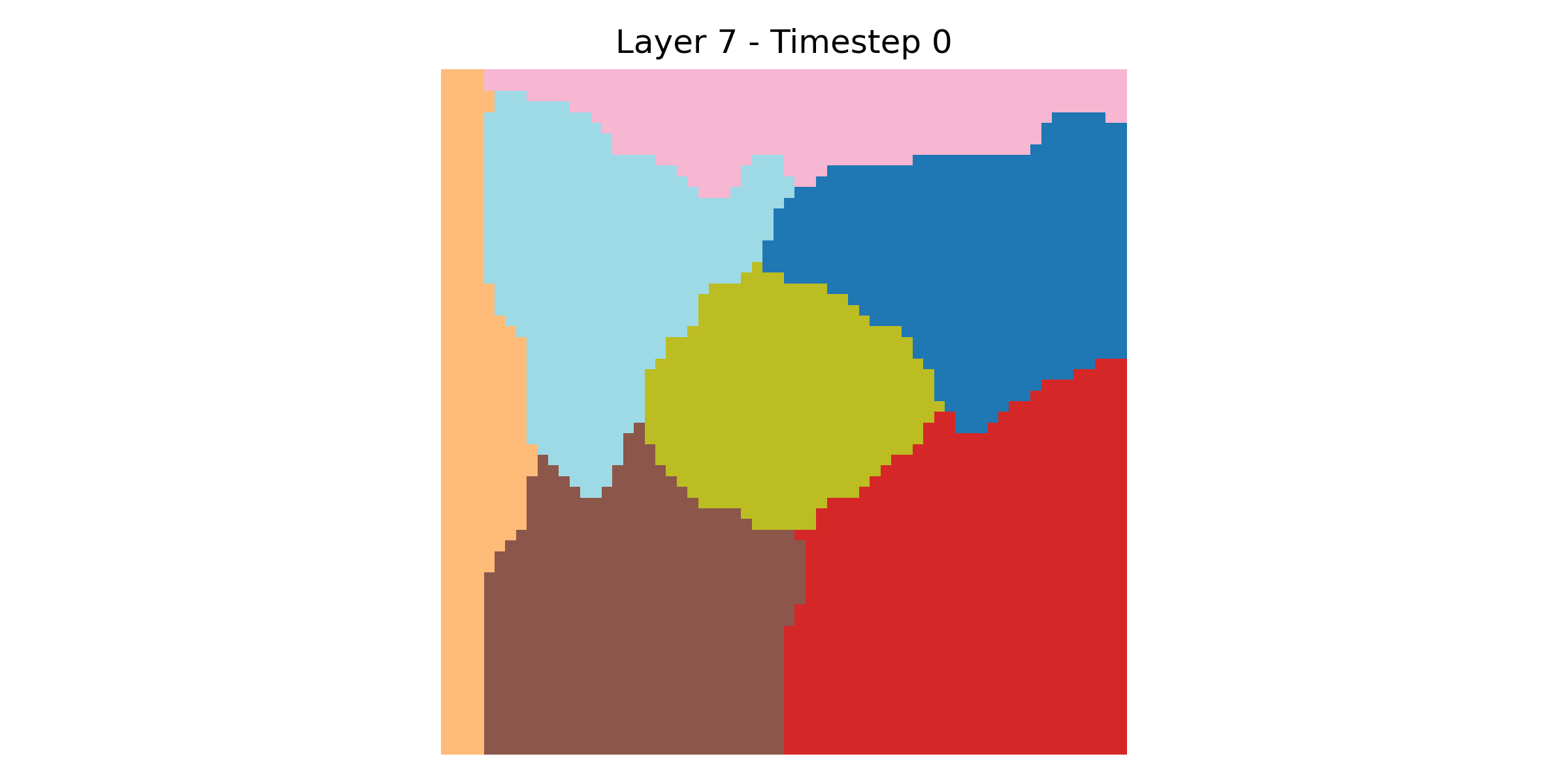} &
\includegraphics[width=0.14\linewidth]{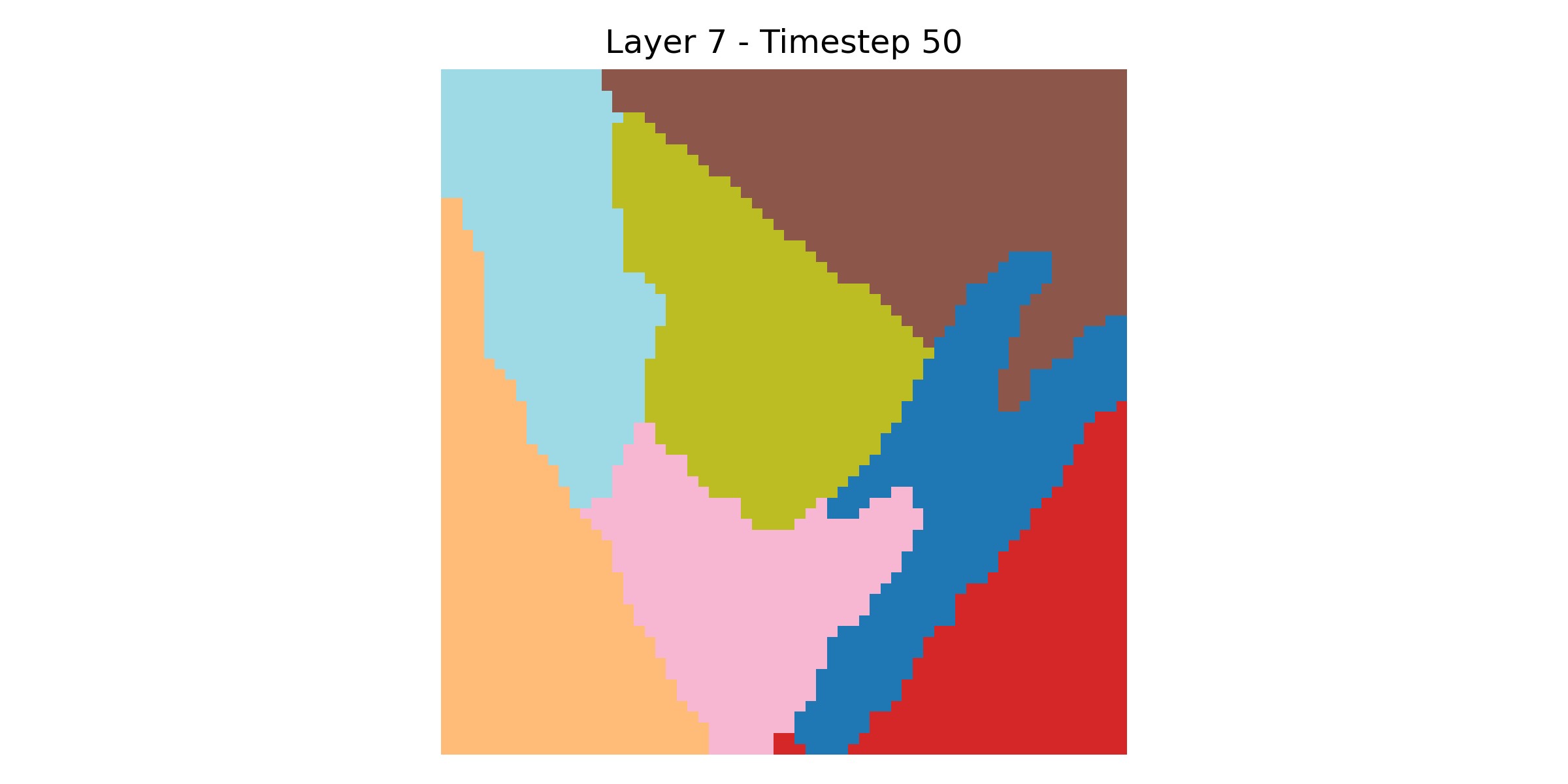} &
\includegraphics[width=0.14\linewidth]{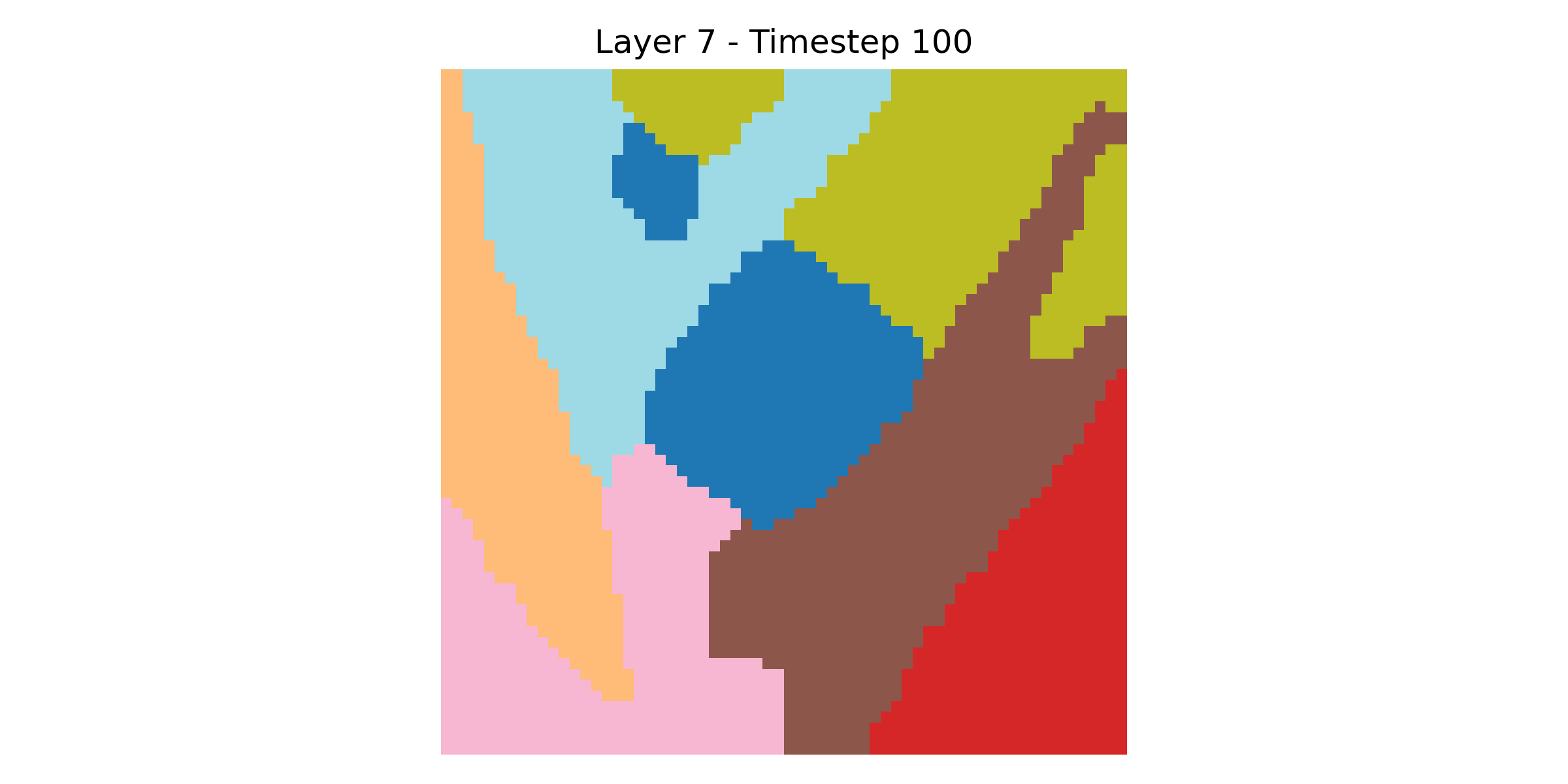} &
\includegraphics[width=0.14\linewidth]{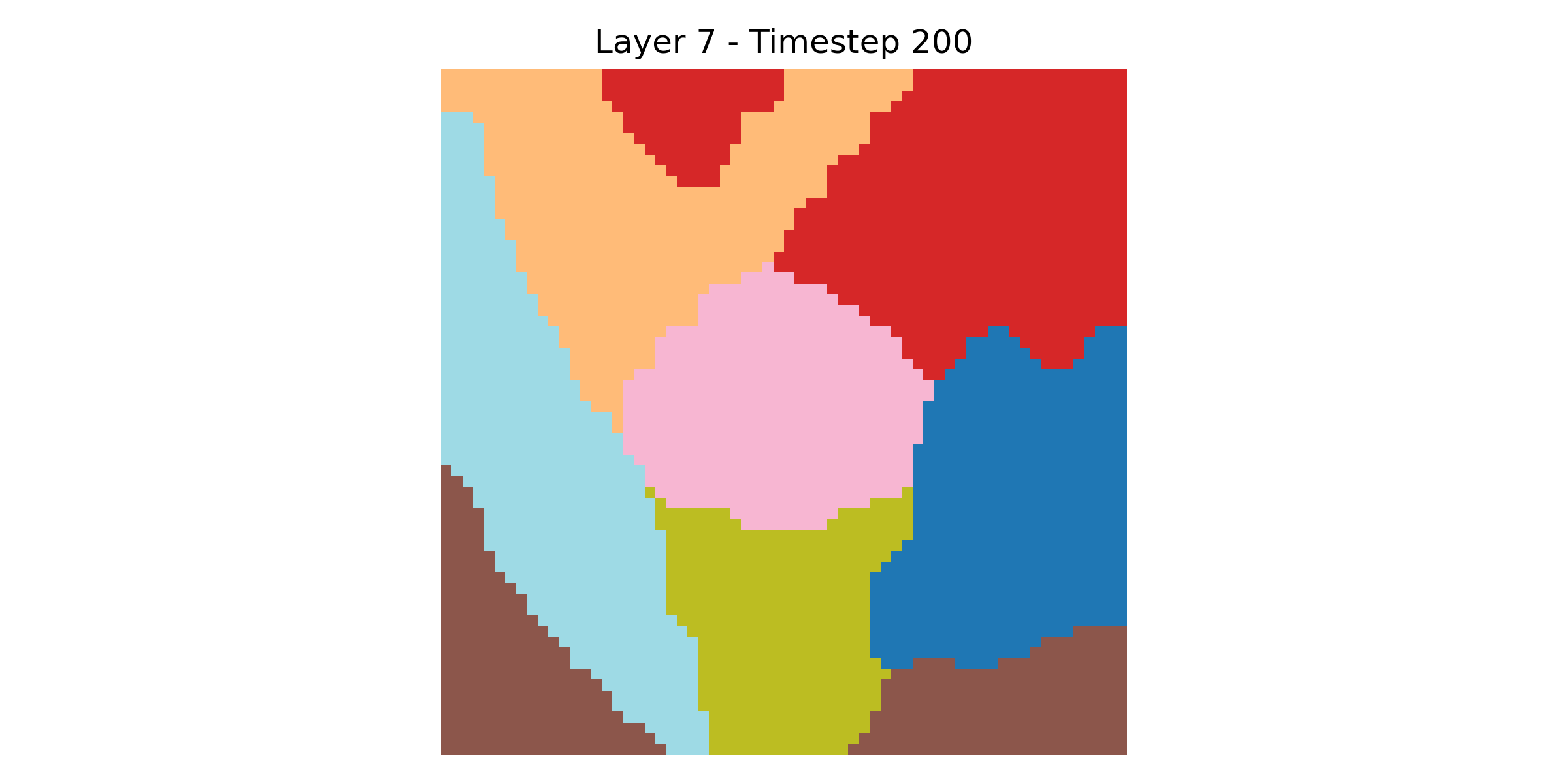} \\

&\raisebox{0.4cm}{Layer 8}  &
\includegraphics[width=0.14\linewidth]{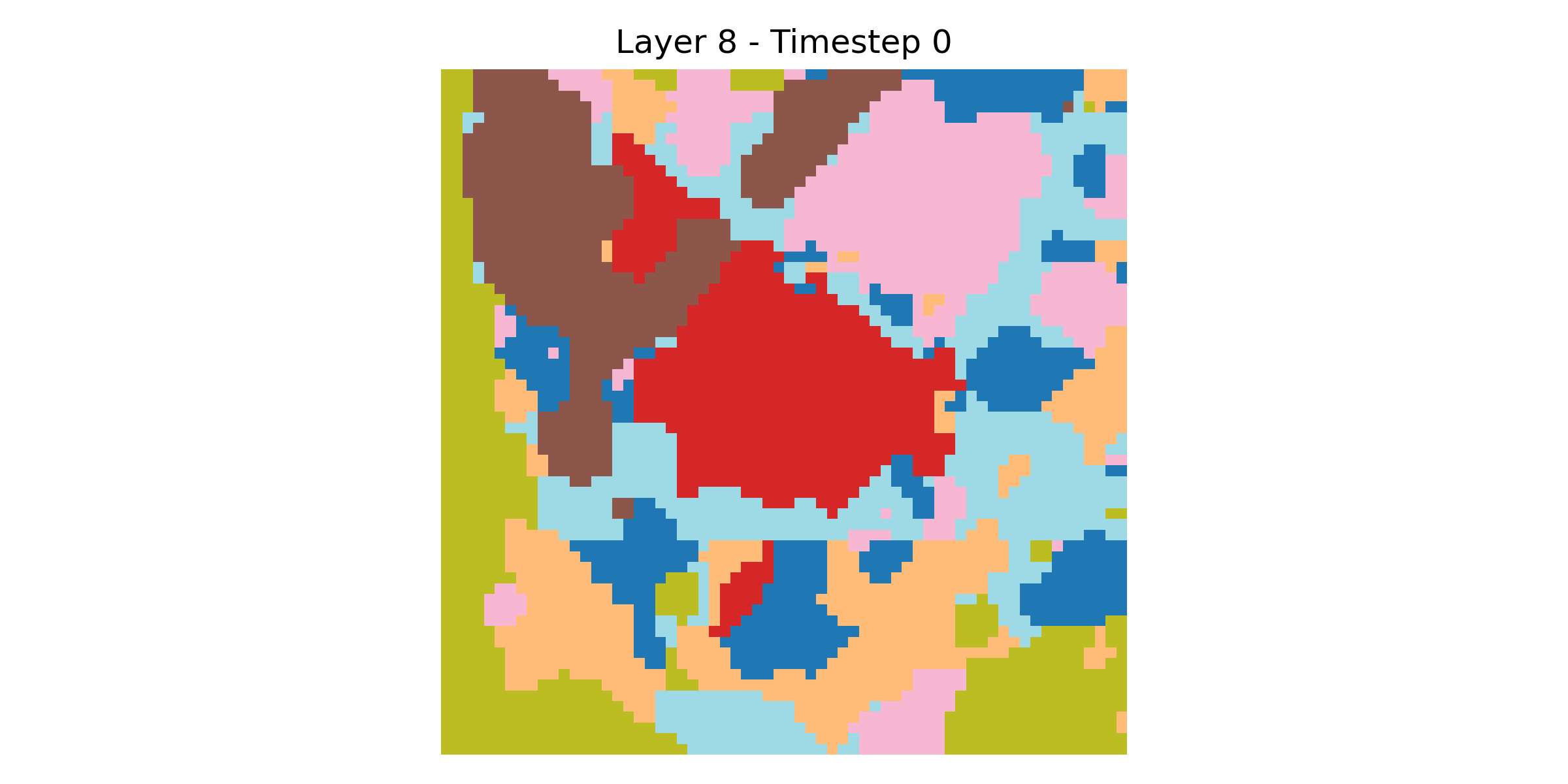} &
\includegraphics[width=0.14\linewidth]{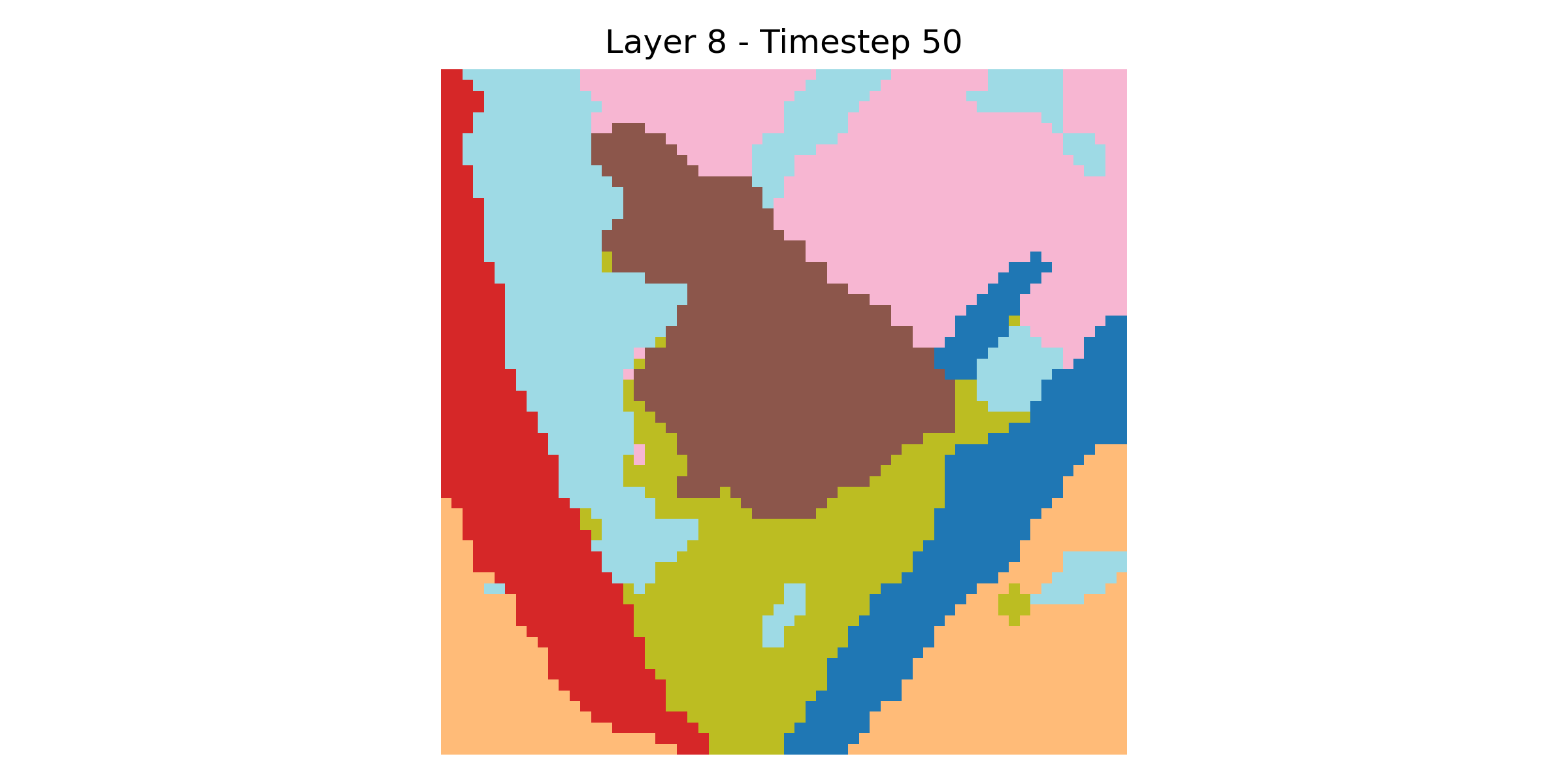} &
\includegraphics[width=0.14\linewidth]{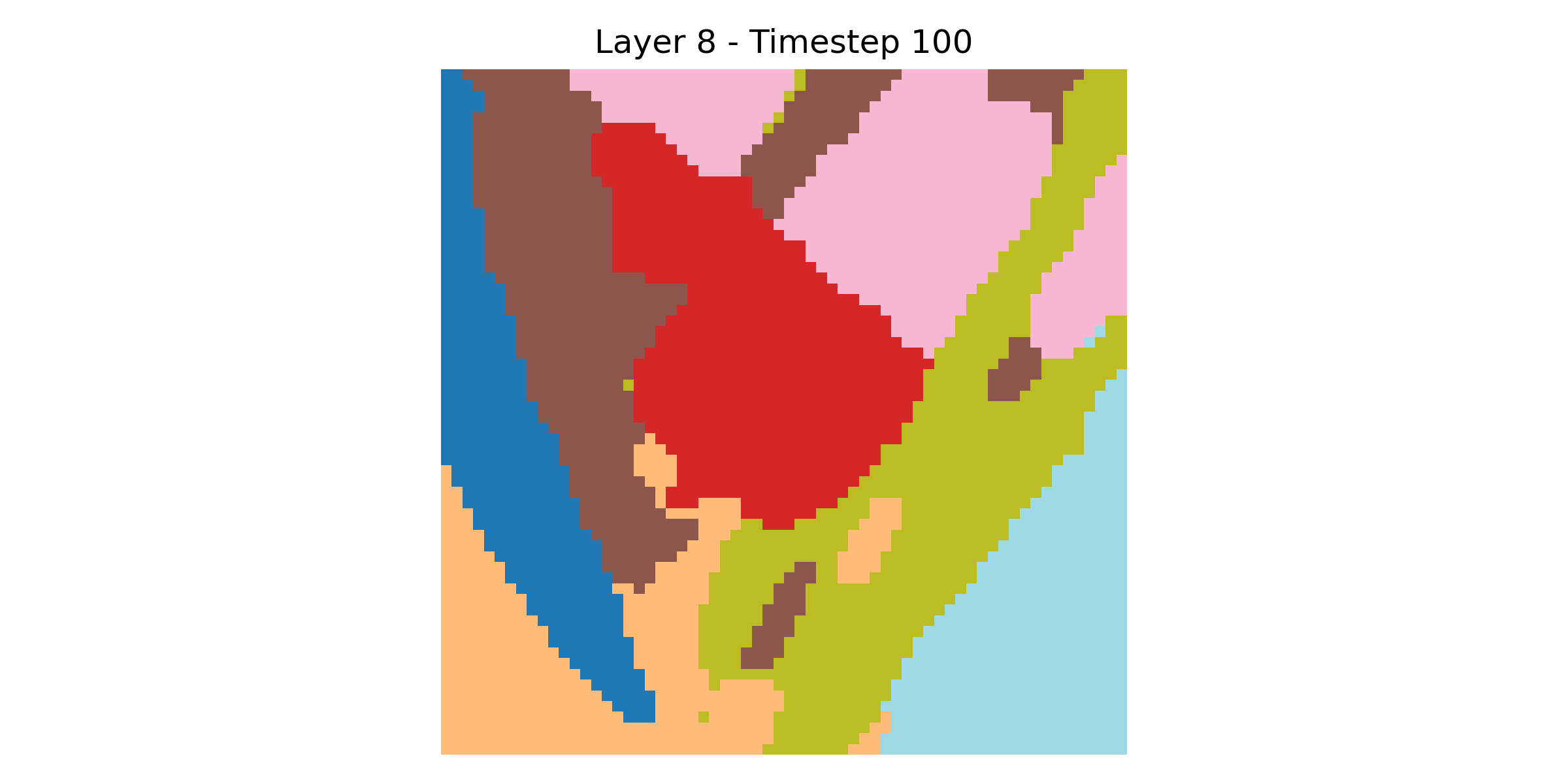} &
\includegraphics[width=0.14\linewidth]{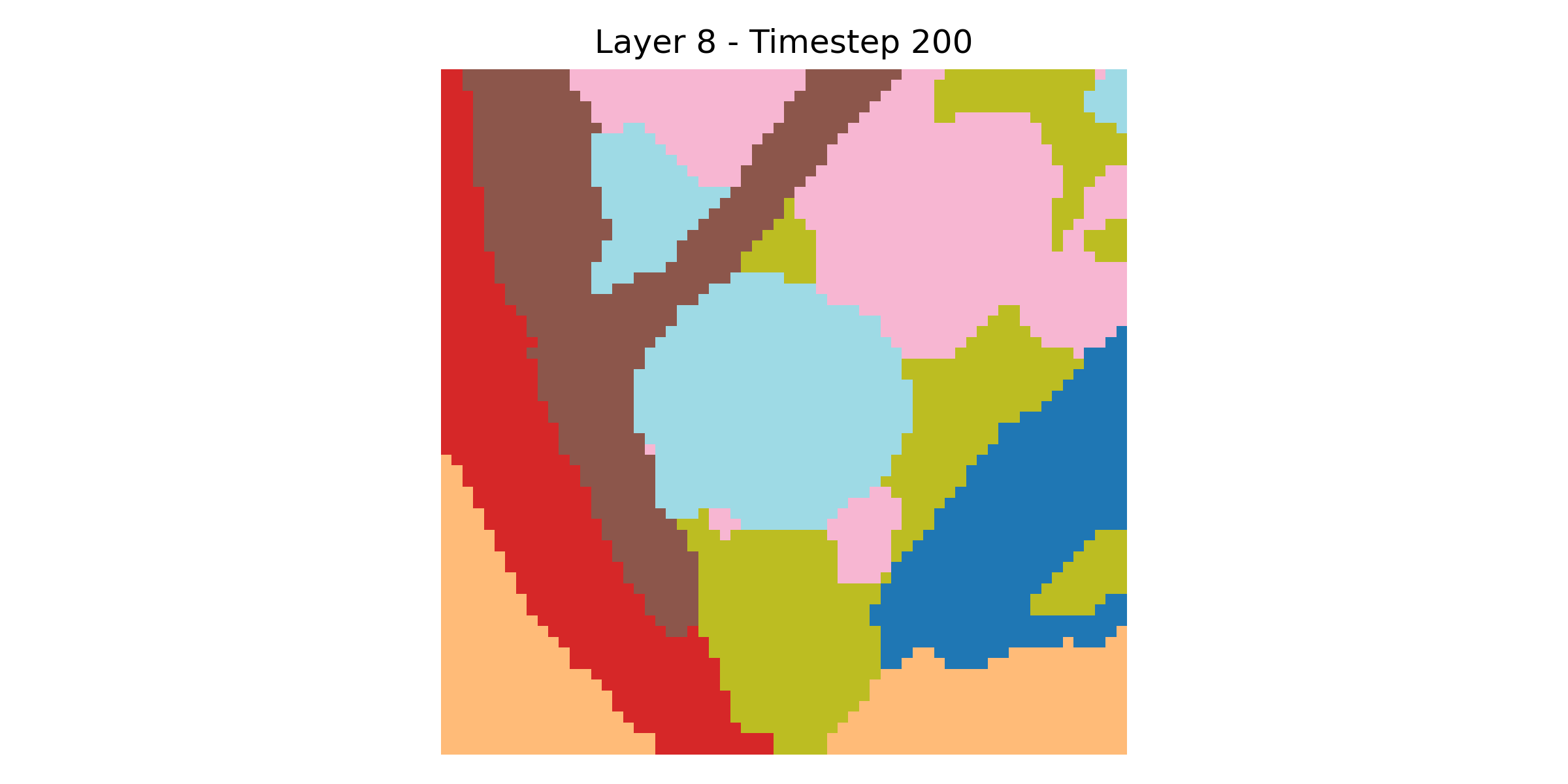} \\

\multirow{3}{*}{
 
    \begin{tabular}{@{}c@{}}
    \includegraphics[height=1.1cm]{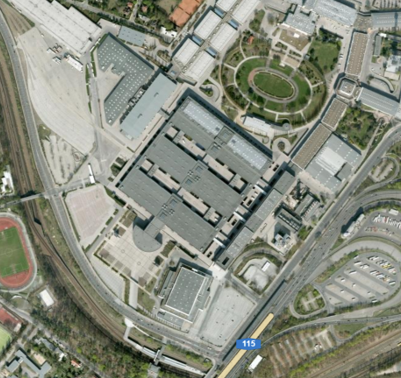} \\
    \scriptsize HS Reference \\ \scriptsize (Google Earth)

    \end{tabular}
  }

& \raisebox{0.4cm}{Layer 9} &

\includegraphics[width=0.14\linewidth]{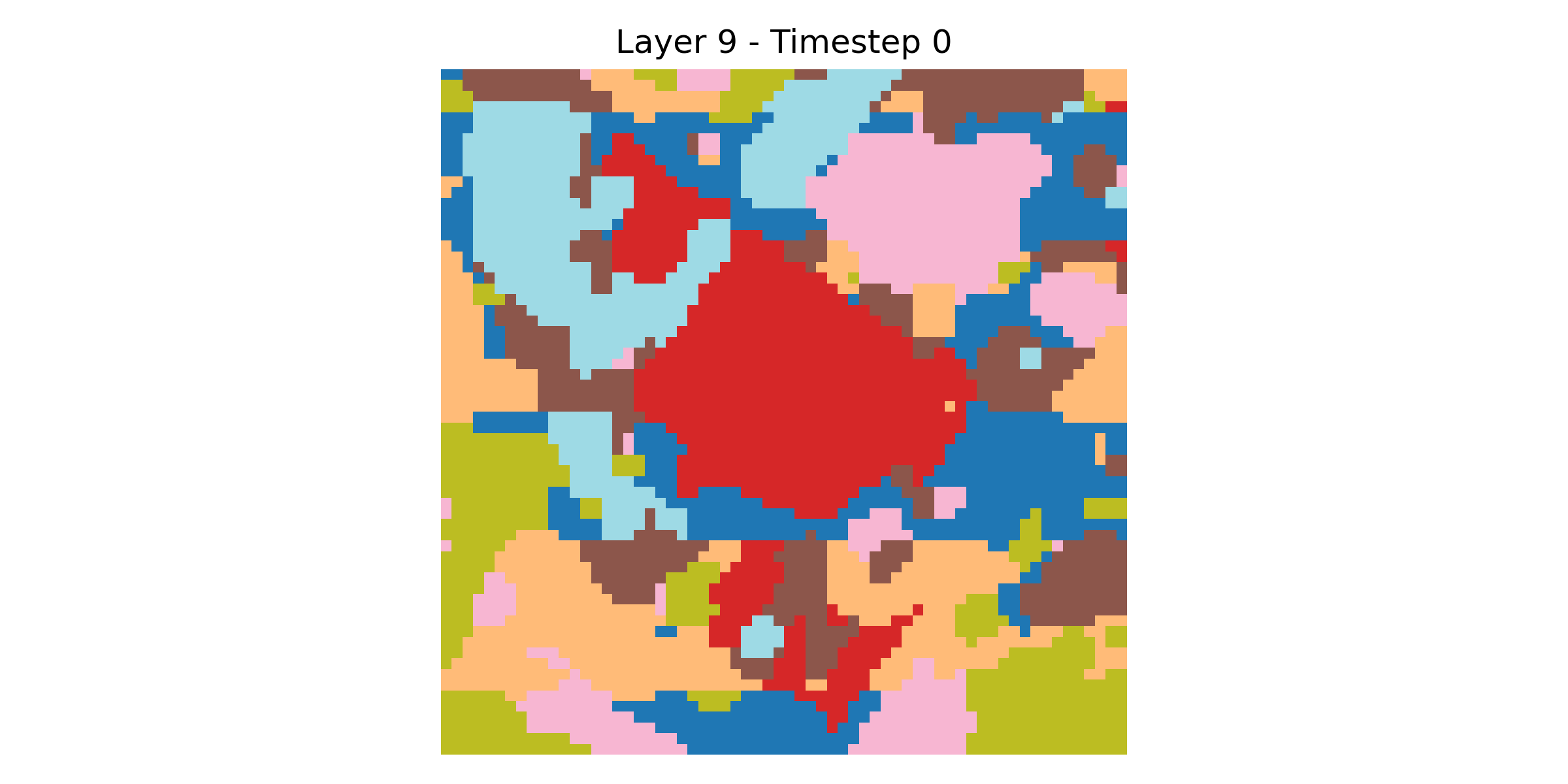} &
\includegraphics[width=0.14\linewidth]{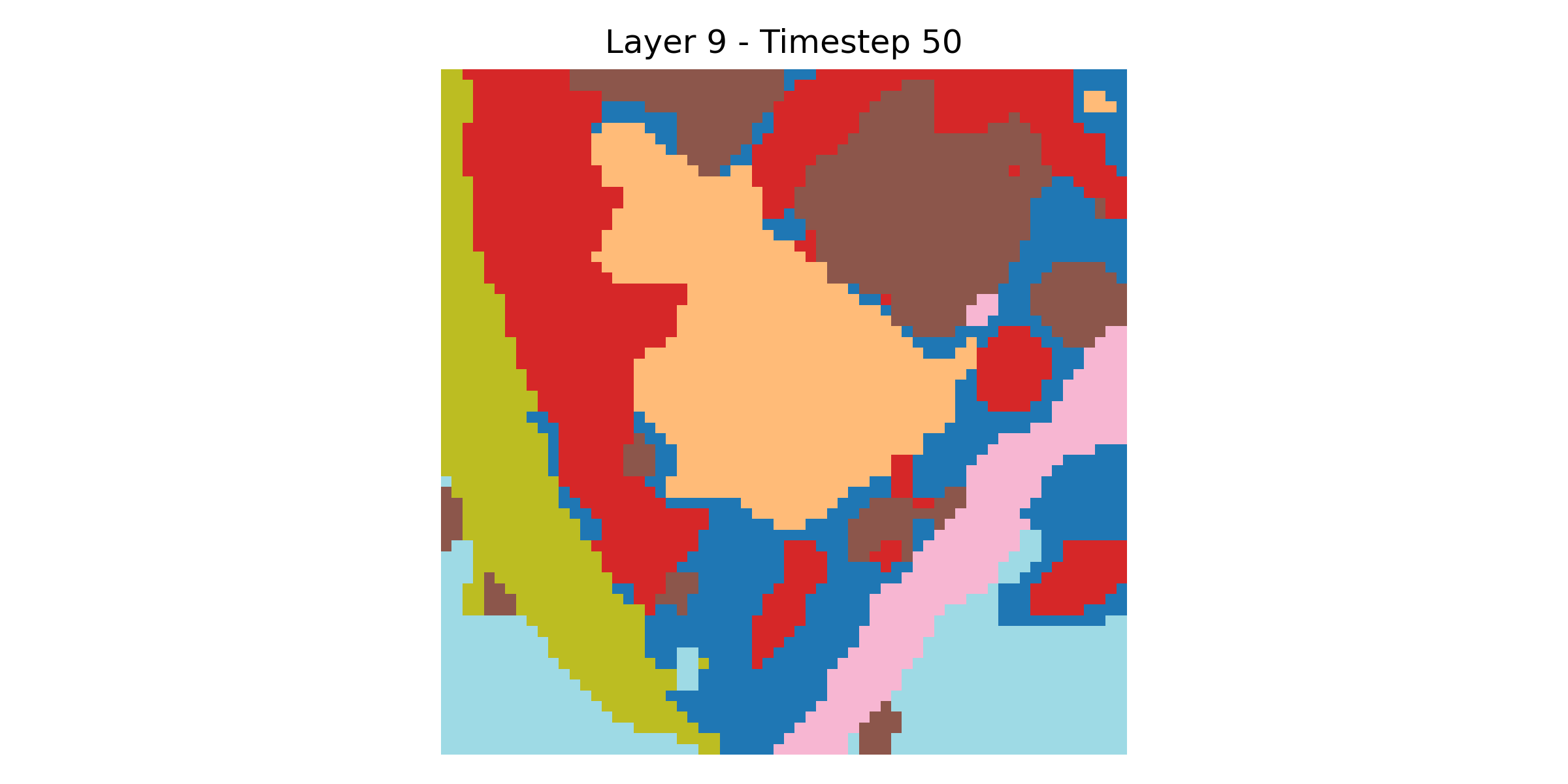} &
\includegraphics[width=0.14\linewidth]{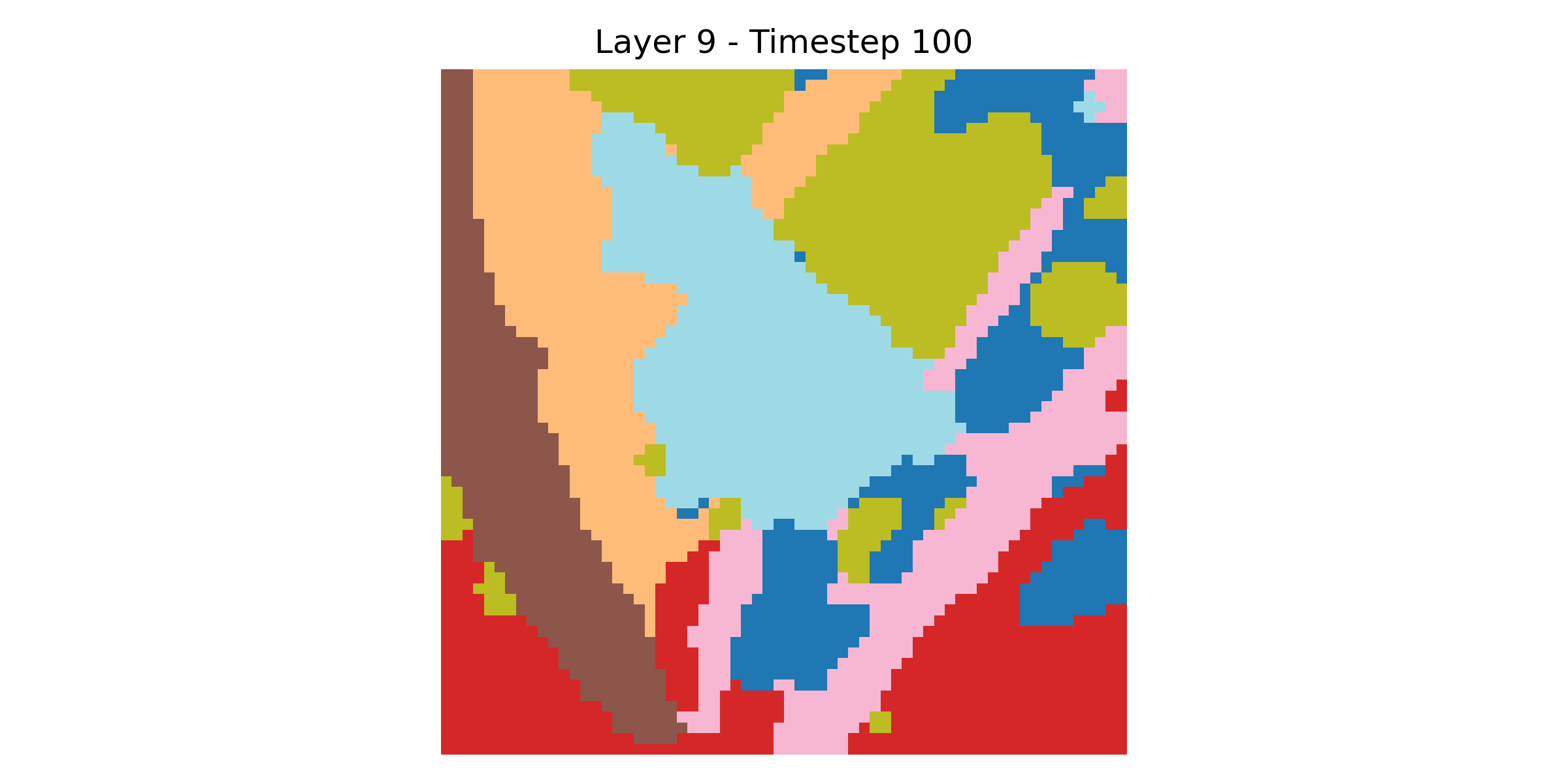} &
\includegraphics[width=0.14\linewidth]{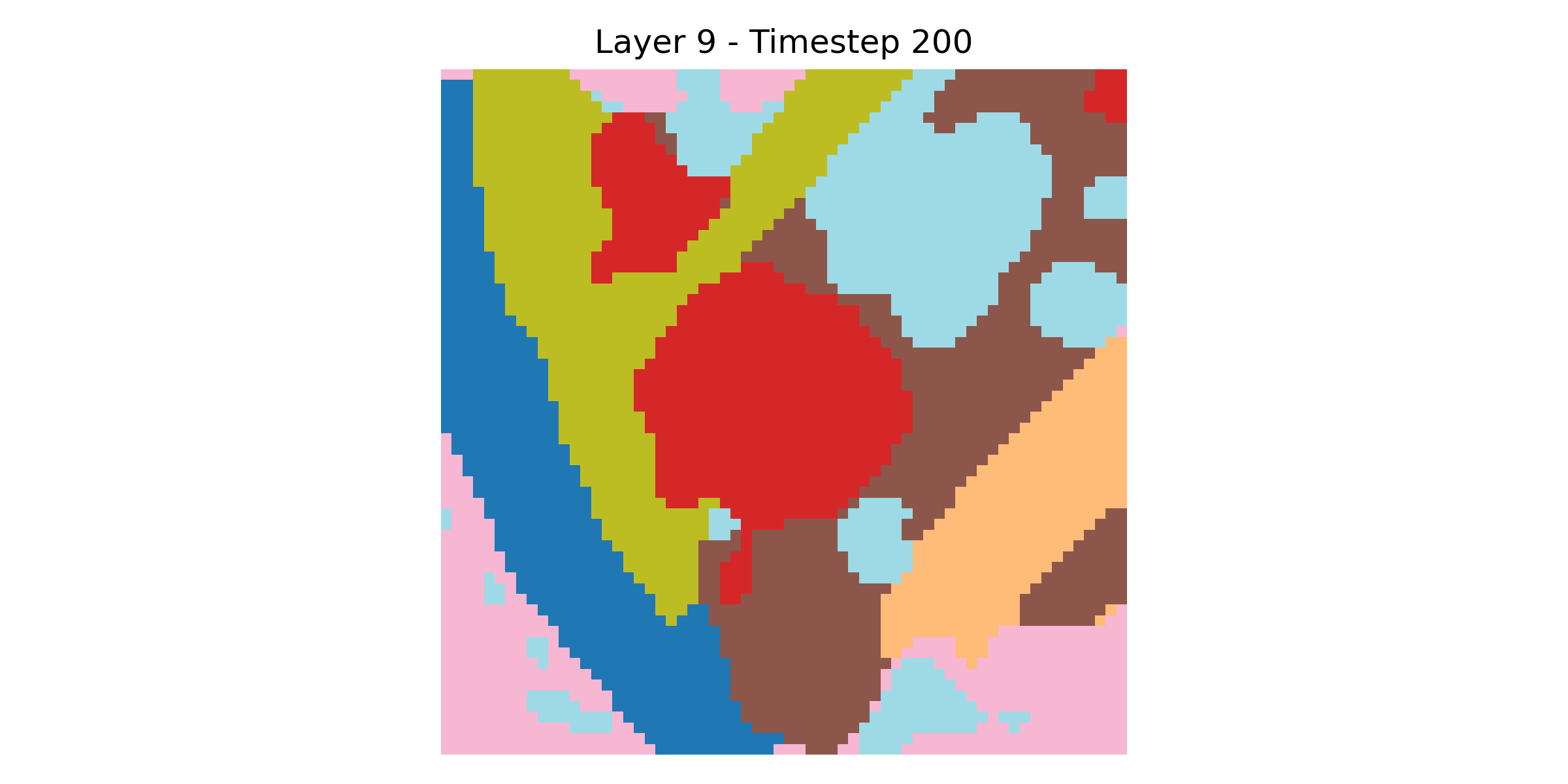} \\

& \raisebox{0.4cm}{Layer 10} &
\includegraphics[width=0.14\linewidth]{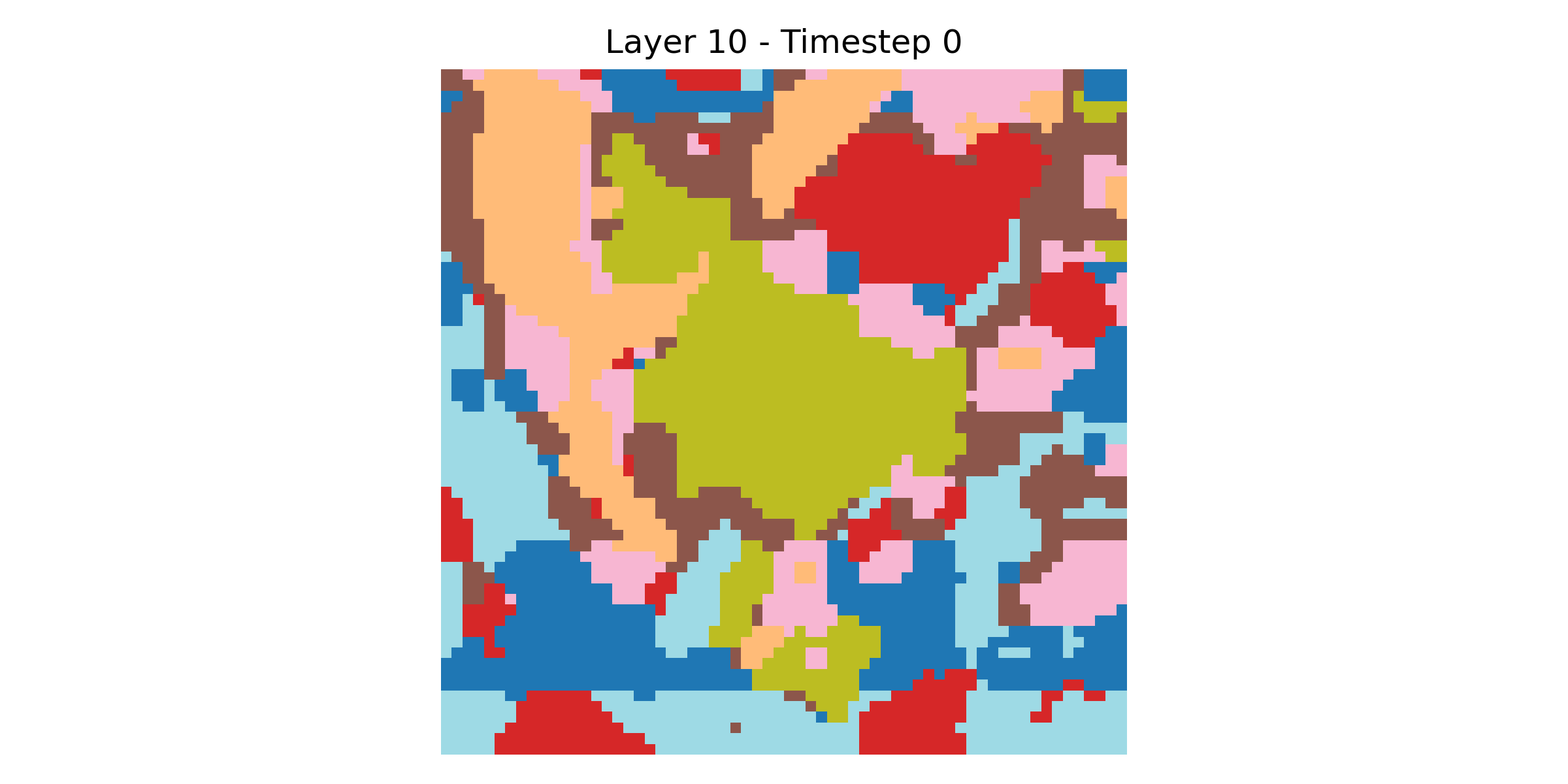} &
\includegraphics[width=0.14\linewidth]{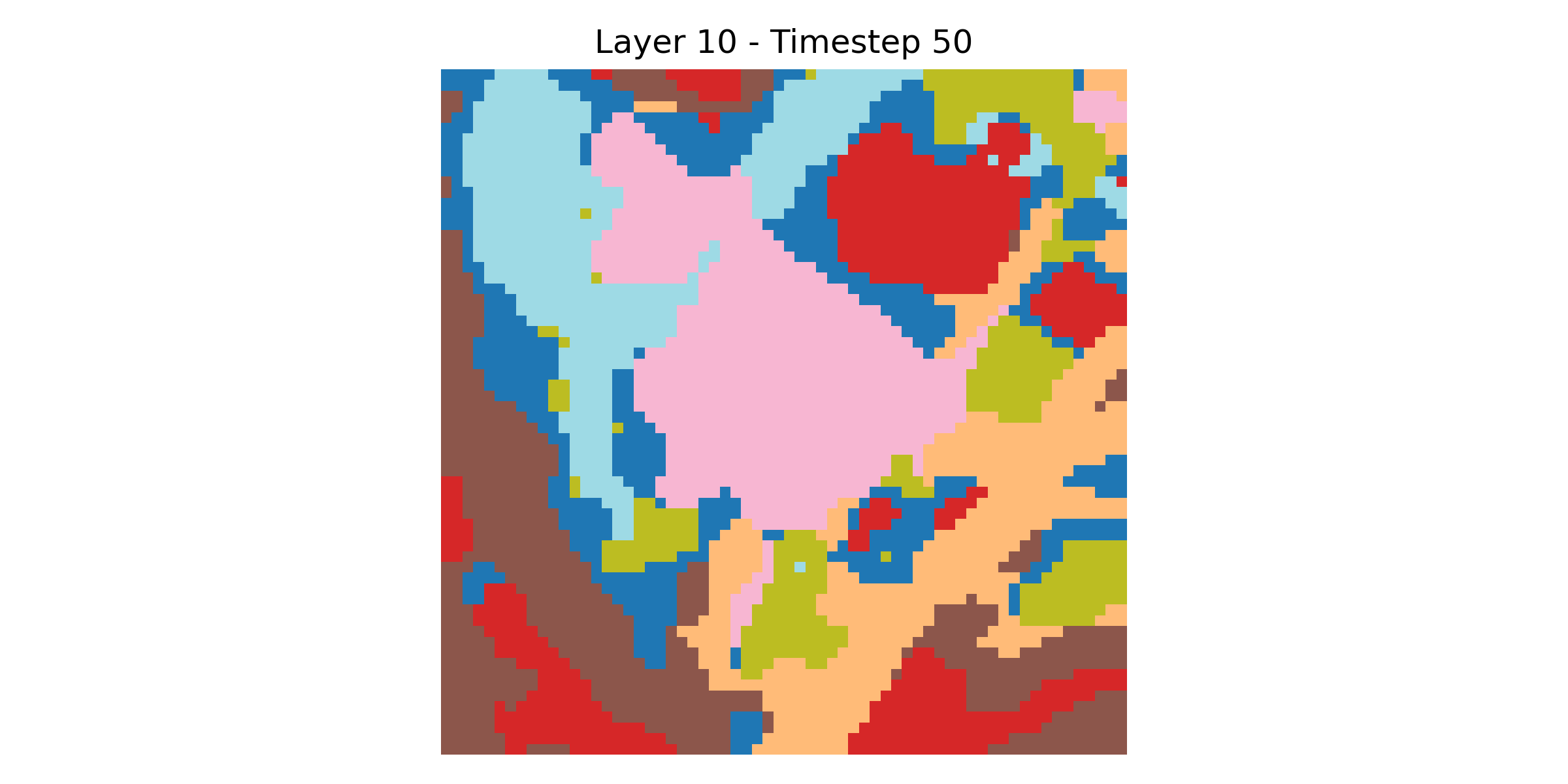} &
\includegraphics[width=0.14\linewidth]{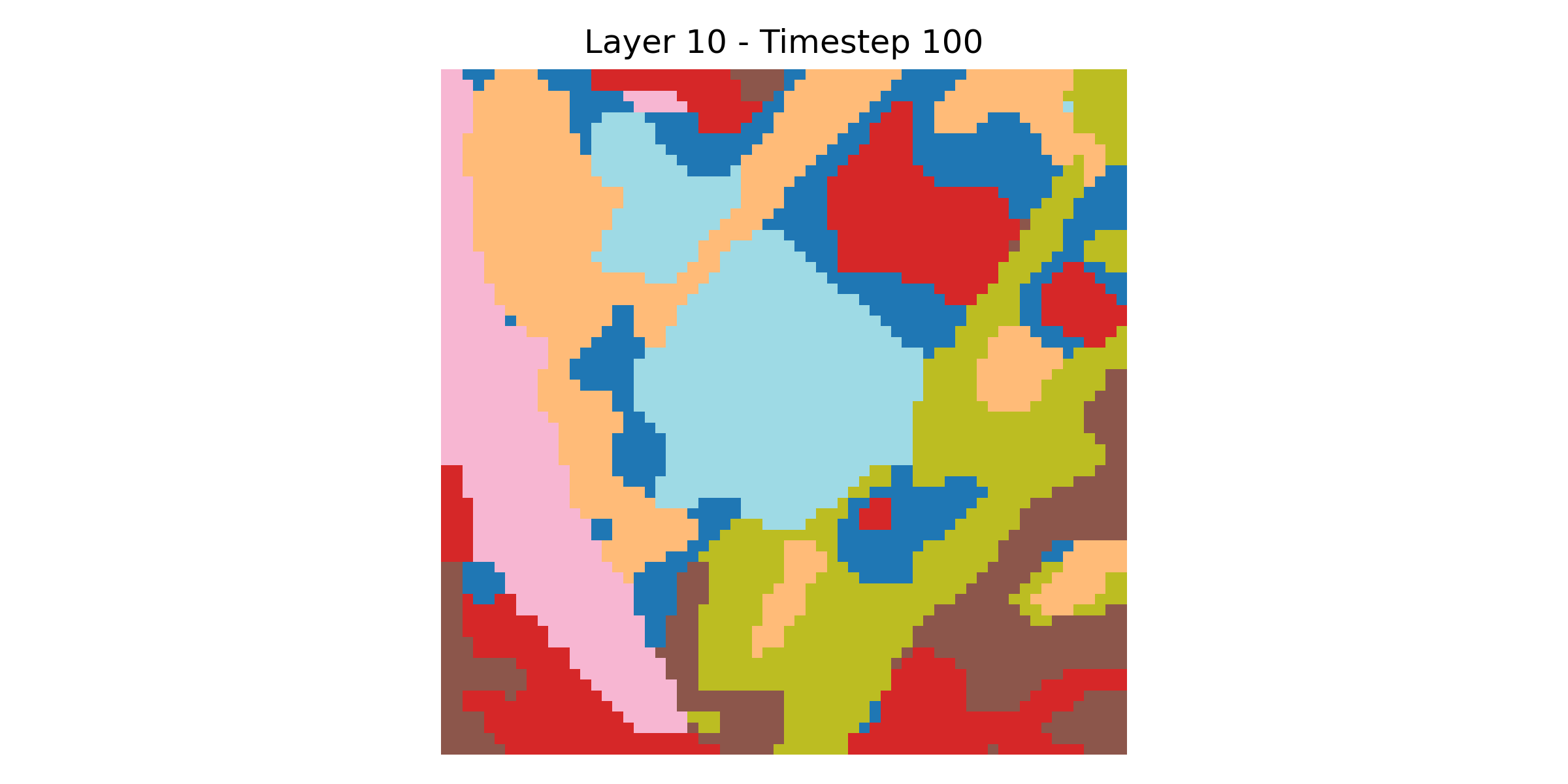} &
\includegraphics[width=0.14\linewidth]{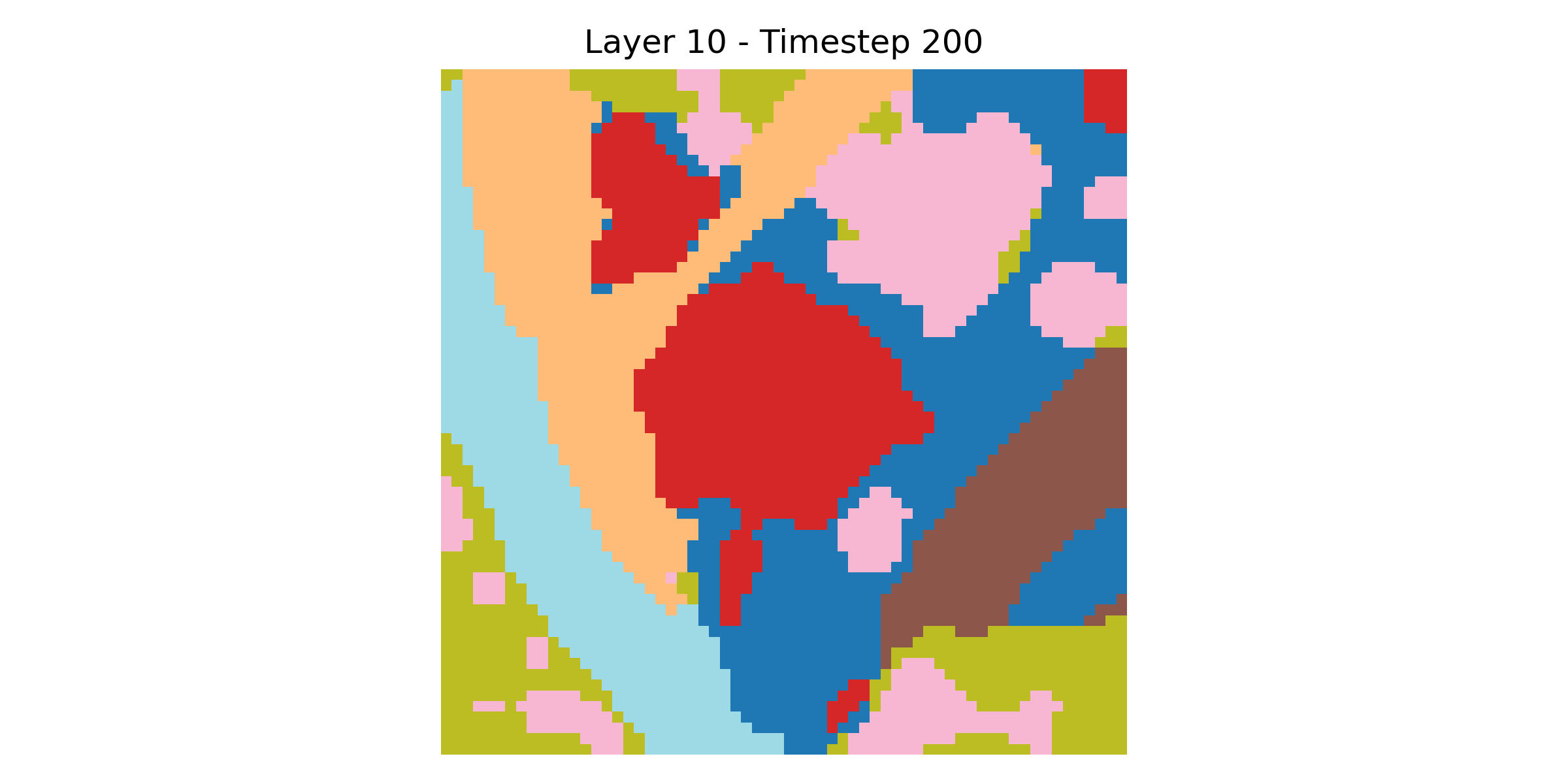} \\

& \raisebox{0.4cm}{Layer 11} &
\includegraphics[width=0.14\linewidth]{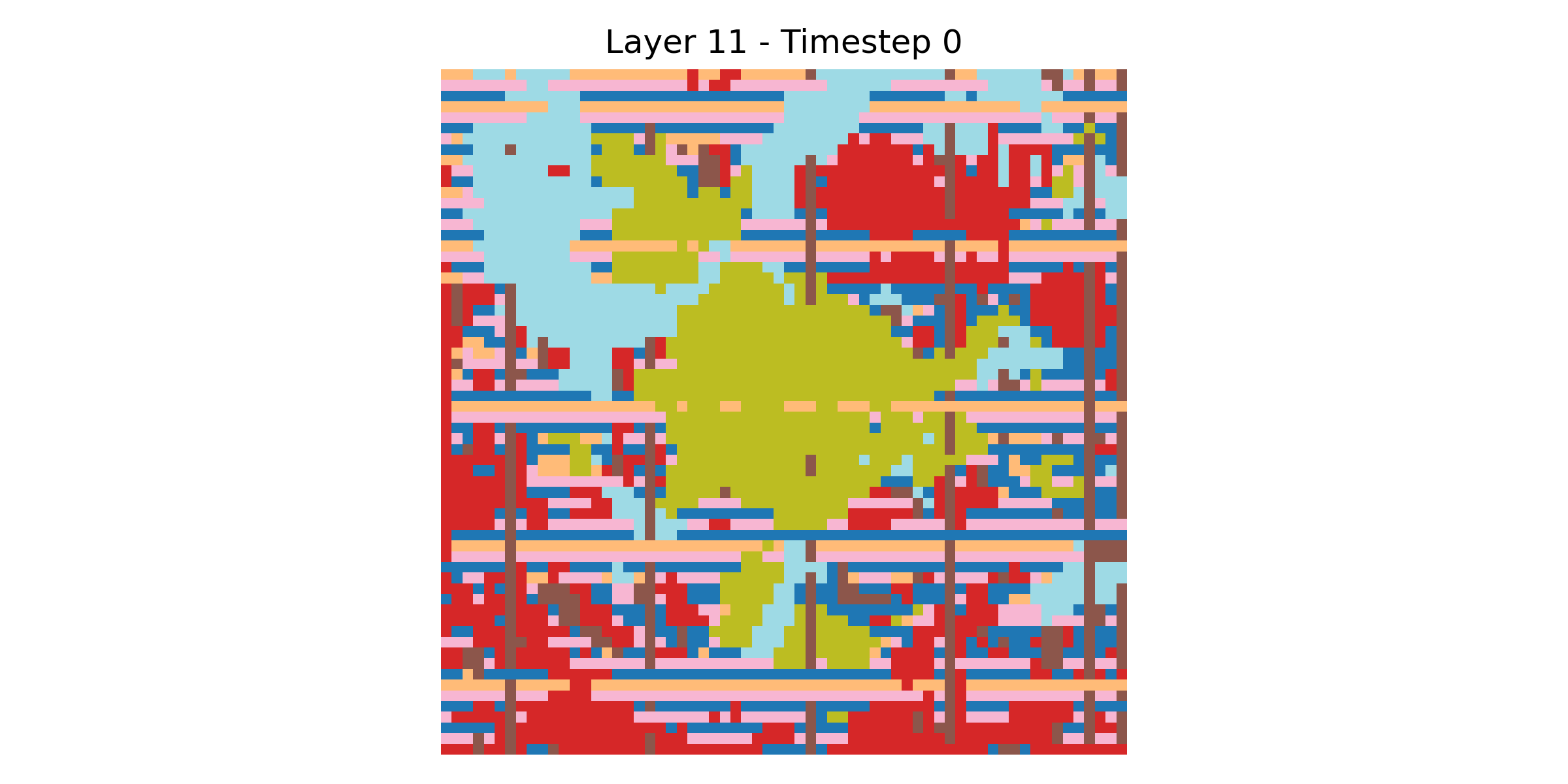} &
\includegraphics[width=0.14\linewidth]{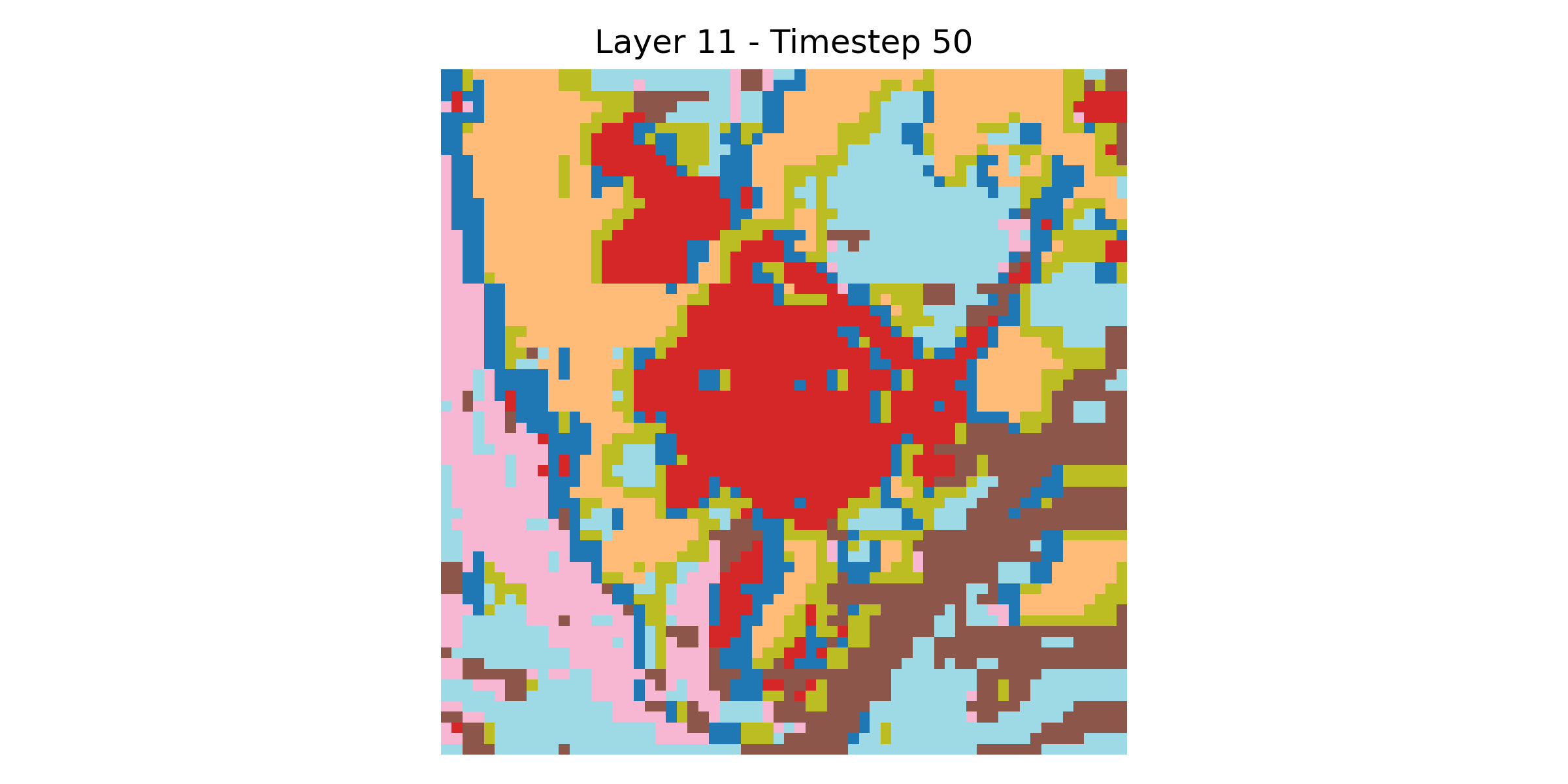} &
\includegraphics[width=0.14\linewidth]{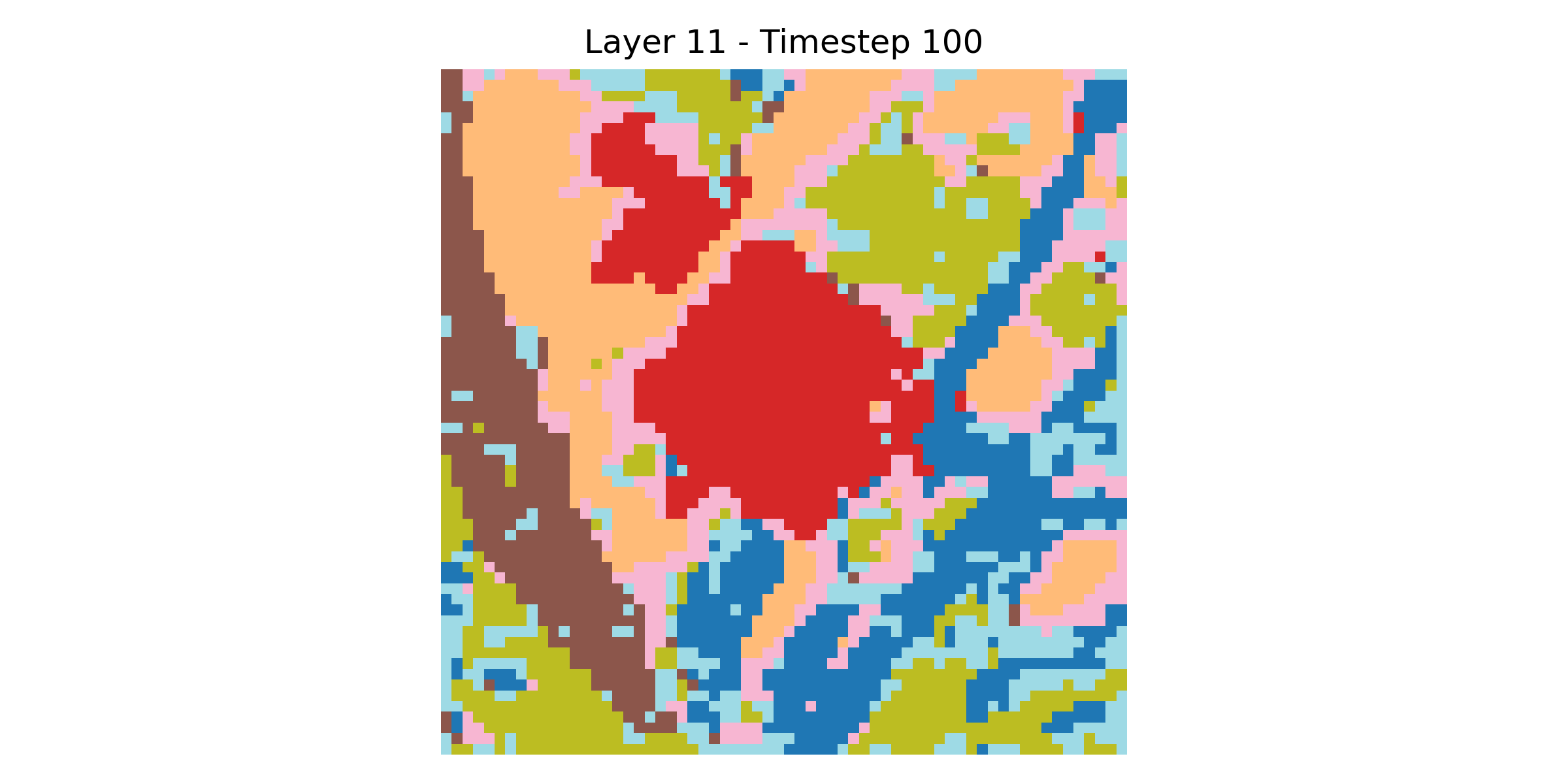} &
\includegraphics[width=0.14\linewidth]{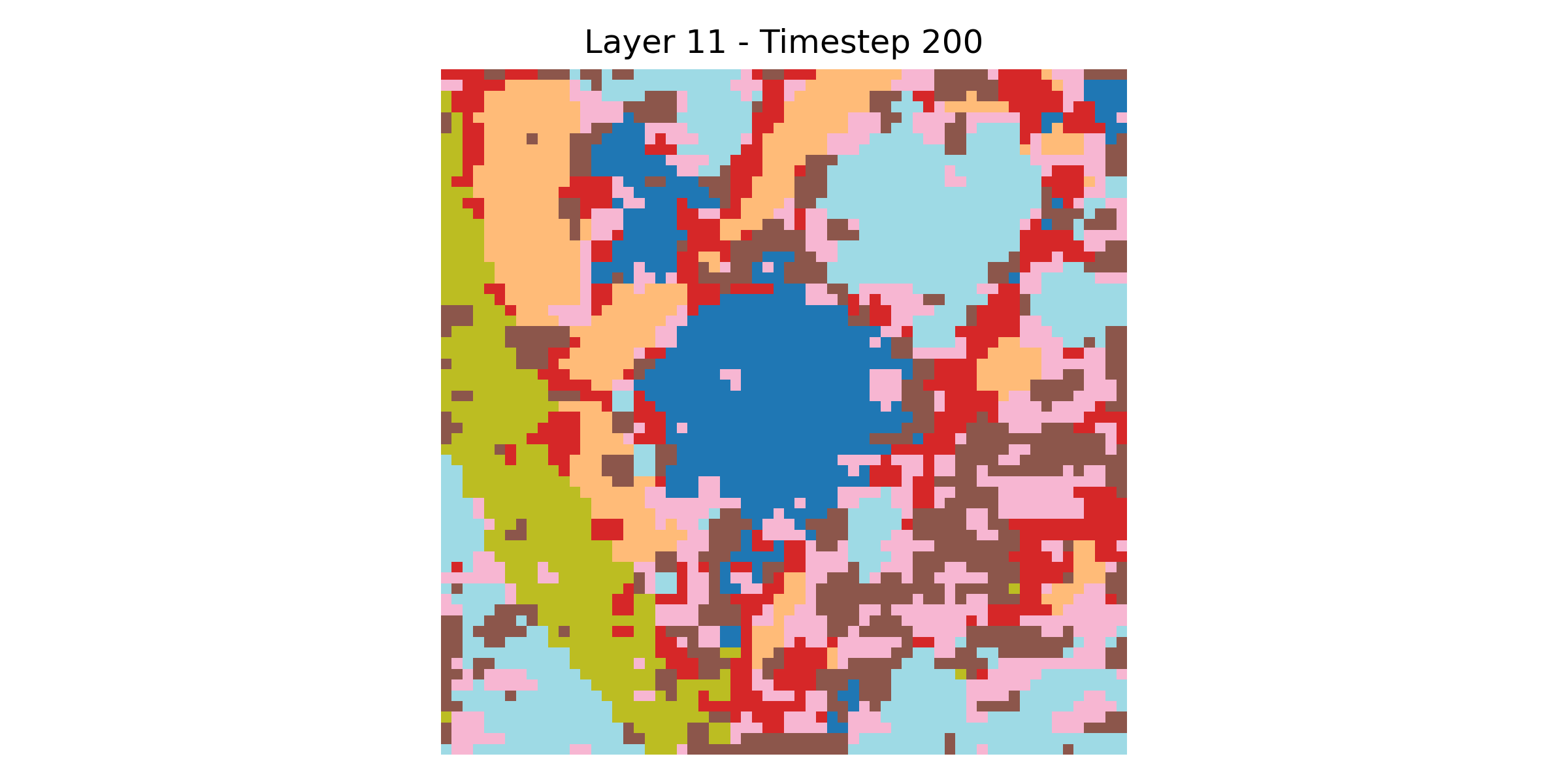} \\

\end{tabular}
\caption{
\textbf{Feature Clustering across decoder layers and timesteps.}
K-means clustering ($k{=}6$) is applied to decoder features from layers 6–11 across timesteps $T{=}0$ to $T{=}200$. The input is a $64{\times}64$ pseudo-RGB patch sampled from the Berlin hyperspectral dataset. 
Left: the original pseudo-RGB patch and a high-resolution reference image from Google Earth (circa 2009) are shown for context. Note that cluster colors are assigned independently in each plot and are therefore not consistent across layers or timesteps.
}
\label{diffuson berlin59 feature visualization}
\end{figure*}

Despite the degraded input quality, the diffusion-derived features produce semantically coherent clusters. Lower and intermediate decoder layers (e.g., Layers 6–7) tend to segment broad, coarse regions, while higher layers (e.g., Layers 9–11) better delineate object boundaries and suppress noise. This progression illustrates a shift from high-level abstraction in earlier layers to more detailed, spatially localized information in later layers.

\section{Detailed Analysis for Dataset Augsburg and Berlin Over Layers}
\label{sec:appendixA}
\subsection{Augsburg Dataset}

\subsubsection{Visualization}
\begin{figure*}[h]
    \centering
    \vskip 0.15in

    \begin{minipage}[b]{0.19\textwidth}
        \includegraphics[width=\linewidth]{figure_Augsburg/test_label.png}
        \centering \textbf{Test label}
    \end{minipage}
    \hfill
    \begin{minipage}[b]{0.19\textwidth}
        \includegraphics[width=\linewidth]{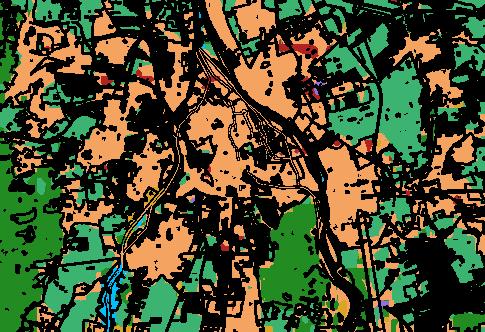}
        \centering \textbf{Layer 2}
    \end{minipage}
    \hfill
    \begin{minipage}[b]{0.19\textwidth}
        \includegraphics[width=\linewidth]{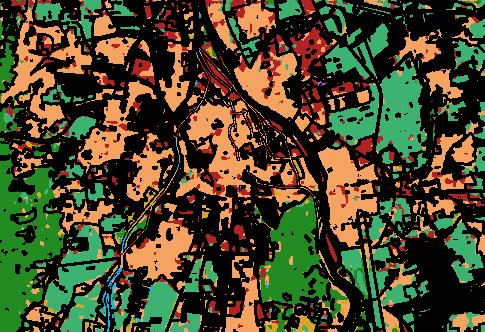}
        \centering \textbf{Layer 5}
    \end{minipage}
    \hfill
    \begin{minipage}[b]{0.19\textwidth}
        \includegraphics[width=\linewidth]{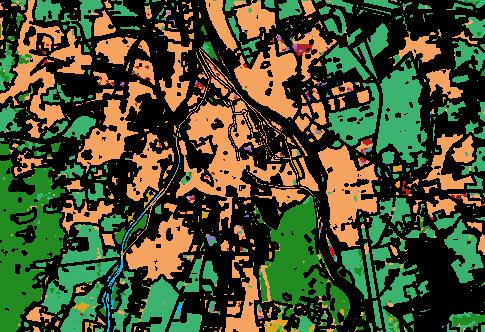}
        \centering \textbf{Layer 8}
    \end{minipage}
    \hfill
    \begin{minipage}[b]{0.19\textwidth}
        \includegraphics[width=\linewidth]{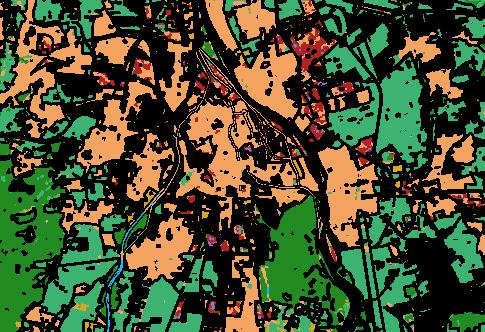}
        \centering \textbf{Layer 10}
    \end{minipage}

    \vskip 0.3em
    \includegraphics[width=\textwidth]{figure_Augsburg/legend_horizontal.pdf}

    \caption{Visualization of the test label and informative layers (Layer 2, Layer 5, Layer 8, Layer 10) from GeoDiffNet with a fixed timestep of 50 for the Augsburg dataset. Higher layers capture more detailed and accurate features.}
    \label{fig:Augsburg layer2-5-8-11}
    \vskip -0.15in
\end{figure*}

In this experiment, we extract spatial features using the diffusion model at different layers and a fixed time-step 0, and evaluate layer informativeness based on performance. As shown in \cref{fig:Augsburg layer2-5-8-11}, the figure illustrates the test labels and the inference (prediction) from the key informative layers (Layers 2, 5, 8, and 10) from the GeoDiffNet model for the Augsburg dataset. By comparing across different layers, we observe that as the layers progress higher, they become more informative. The progression through the layers demonstrates an increasing level of detail and accuracy in feature capture. Specifically, higher layers (Layer 10) exhibit more refined and precise feature representations, resulting in clearer delineations and more accurate classifications compared to the coarser and less detailed representations in the lower layers.

\begin{table*}[h]
  \caption{Performance for U-Net layers in the diffusion model with spatial features at \textbf{timestep 0} for the \textbf{Augsburg} dataset.}
  \centering
  \setlength{\tabcolsep}{2.5pt}  
  \renewcommand{\arraystretch}{1.0}  
  {\small
  \begin{tabular}{@{}lcccccccccccc@{}}
    \toprule
    & \multicolumn{12}{c}{Layers} \\
    \cmidrule(lr){2-13}
    & Layer 1 & Layer 2 & Layer 3 & Layer 4 & Layer 5 & Layer 6 & Layer 7 & Layer 8 & Layer 9 & Layer 10 & Layer 11 & Layer 12 \\
    \midrule
    \textbf{Forest}            & 78.29 & 83.17 & 80.12 & 83.47 & 80.52 & 89.14 & 83.23 & 90.49 & 91.62 & 92.78 & 92.70 & 91.98 \\
    \textbf{Residential}       & 95.36 & 96.69 & 97.16 & 98.49 & 89.51 & 98.39 & 98.92 & 98.59 & 98.25 & 98.04 & 98.12 & 91.00 \\
    \textbf{Industrial}        & 14.78 & 15.46 & 16.89 & 18.33 & 79.06 & 25.64 & 12.53 & 15.74 & 35.69 & 61.46 & 46.06 & 73.13 \\
    \textbf{Low Plants}        & 73.70 & 79.94 & 86.42 & 83.54 & 74.97 & 86.72 & 91.01 & 91.12 & 93.80 & 95.98 & 93.25 & 91.52 \\
    \textbf{Allotment}         & 28.30 & 26.00 & 29.45 & 36.71 & 57.17 & 57.17 & 47.23 & 70.36 & 75.72 & 86.42 & 78.97 & 55.64 \\
    \textbf{Commercial}        & 1.71  & 2.08  & 1.28  & 0.98  & 0.98  & 4.21  & 2.93  & 8.30  & 8.85  & 6.35  & 4.27  & 3.60  \\
    \textbf{Water}             & 10.35 & 10.22 & 11.15 & 16.39 & 11.41 & 16.39 & 15.66 & 17.85 & 18.91 & 14.33 & 14.80 & 13.07 \\
    \midrule
    \textbf{Overall Accuracy (\%)}  & 76.92 & 80.44 & 82.42 & 82.74 & 78.85 & 85.34 & 85.24 & 86.87 & 88.91 & 90.98 & 89.21 & 86.86 \\
    \textbf{AA (\%)}                & 43.21 & 44.79 & 46.07 & 48.27 & 56.23 & 53.95 & 50.22 & 56.06 & 60.41 & 65.05 & 61.17 & 59.99 \\
    \textbf{Kappa Coefficient}      & 0.6552 & 0.7096 & 0.7381 & 0.7434 & 0.7023 & 0.7836 & 0.7792 & 0.8059 & 0.8368 & 0.8682 & 0.8424 & 0.8125 \\
    \bottomrule
  \end{tabular}
  }

  \label{tab:layer_Augsburg_time0}
\end{table*}

\begin{table*}[h]

  \caption{Performance for U-Net layers in the diffusion model with spatial features at \textbf{timestep 50} for the \textbf{Augsburg} dataset.}
  \centering
  \setlength{\tabcolsep}{2.5pt}  
  \renewcommand{\arraystretch}{1.0}  

  {\small
  \begin{tabular}{@{}lcccccccccccc@{}}
    \toprule
    & \multicolumn{12}{c}{Layers} \\
    \cmidrule(lr){2-13}
    & Layer 1 & Layer 2 & Layer 3 & Layer 4 & Layer 5 & Layer 6 & Layer 7 & Layer 8 & Layer 9 & Layer 10 & Layer 11 & Layer 12 \\
    \midrule
    \textbf{Forest}            & 72.01 & 71.51 & 79.22 & 78.01 & 81.18 & 68.75 & 86.63 & 91.53 & 90.40 & 90.88 & 89.84 & 89.41 \\
    \textbf{Residential}       & 91.73 & 94.53 & 95.77 & 95.42 & 96.27 & 95.99 & 97.79 & 97.36 & 97.00 & 97.16 & 92.93 & 97.44 \\
    \textbf{Industrial}        & 6.76  & 4.23  & 12.09 & 5.27  & 10.23 & 7.18  & 9.95  & 16.53 & 26.34 & 25.56 & 49.19 & 31.70 \\
    \textbf{Low Plants}        & 58.86 & 67.56 & 70.32 & 74.74 & 74.21 & 79.45 & 77.61 & 78.42 & 82.91 & 81.85 & 76.92 & 75.49 \\
    \textbf{Allotment}         & 10.71 & 6.50  & 13.96 & 21.99 & 26.58 & 23.90 & 18.16 & 24.67 & 30.40 & 39.77 & 46.08 & 24.47 \\
    \textbf{Commercial}        & 0.12  & 0.00  & 0.73  & 0.98  & 0.92  & 1.04  & 0.73  & 3.17  & 4.27  & 4.33  & 4.40  & 2.56  \\
    \textbf{Water}             & 4.58  & 10.29 & 9.29  & 13.87 & 15.20 & 9.22  & 16.99 & 12.08 & 14.73 & 14.47 & 9.36  & 10.22 \\
    \midrule
    \textbf{Overall Accuracy (\%)}         & 68.68 & 72.62 & 75.81 & 76.79 & 77.79 & 77.05 & 80.44 & 81.72 & 83.53 & 83.33 & 80.93 & 81.08 \\
    \textbf{AA (\%)}                       & 34.97 & 36.37 & 40.20 & 41.47 & 43.51 & 40.79 & 43.98 & 46.25 & 49.44 & 50.58 & 52.67 & 47.33 \\
    \textbf{Kappa Coefficient}             & 0.526 & 0.585 & 0.639 & 0.651 & 0.667 & 0.653 & 0.708 & 0.730 & 0.757 & 0.755 & 0.727 & 0.723 \\
    \bottomrule
  \end{tabular}
  }

  \label{tab:layer_Augsburg_timestep50}
\end{table*}

\begin{table*}[h]
  \caption{Performance for U-Net layers in the diffusion model with spatial features at \textbf{timestep 100} for the \textbf{Augsburg} dataset.}
  \centering
  \setlength{\tabcolsep}{2.5pt}  
  \renewcommand{\arraystretch}{1.0}  
  {\small
  \begin{tabular}{@{}lcccccccccccc@{}}
    \toprule
    & \multicolumn{12}{c}{Layers} \\
    \cmidrule(lr){2-13}
    & Layer 1 & Layer 2 & Layer 3 & Layer 4 & Layer 5 & Layer 6 & Layer 7 & Layer 8 & Layer 9 & Layer 10 & Layer 11 & Layer 12 \\
    \midrule
    \textbf{Forest}            & 68.48 & 72.04 & 81.40 & 66.32 & 80.56 & 87.77 & 91.22 & 91.39 & 92.15 & 88.45 & 90.21 & 83.80 \\
    \textbf{Residential}       & 94.22 & 92.78 & 94.63 & 90.25 & 94.20 & 93.99 & 79.94 & 95.45 & 94.78 & 96.05 & 94.90 & 94.77 \\
    \textbf{Industrial}        & 2.66  & 3.16  & 5.17  & 4.80  & 6.03  & 13.89 & 47.68 & 19.19 & 19.95 & 26.61 & 42.56 & 36.53 \\
    \textbf{Low Plants}        & 46.79 & 50.92 & 56.24 & 58.36 & 61.84 & 71.91 & 60.70 & 67.32 & 76.30 & 77.09 & 75.70 & 70.79 \\
    \textbf{Allotment}         & 8.80  & 5.93  & 3.63  & 19.69 & 19.31 & 21.41 & 20.08 & 25.24 & 26.58 & 26.96 & 22.18 & 23.71 \\
    \textbf{Commercial}        & 0.55  & 0.00  & 0.06  & 2.75  & 0.98  & 1.40  & 0.67  & 0.67  & 1.53  & 3.72  & 4.82  & 4.15  \\
    \textbf{Water}             & 1.66  & 4.11  & 8.10  & 6.04  & 6.50  & 12.08 & 11.75 & 12.21 & 10.82 & 11.94 & 11.15 & 9.82 \\
    \midrule
    \textbf{Overall Accuracy (\%)}         & 64.63 & 66.14 & 70.47 & 67.00 & 72.21 & 77.35 & 70.28 & 77.24 & 80.23 & 80.75 & 80.90 & 77.73 \\
    \textbf{AA (\%)}                       & 31.88 & 32.71 & 35.61 & 35.46 & 38.49 & 43.21 & 44.58 & 44.50 & 46.01 & 47.26 & 48.79 & 46.22 \\
    \textbf{Kappa Coefficient}             & 0.4600 & 0.4875 & 0.5578 & 0.5047 & 0.5830 & 0.6644 & 0.5823 & 0.6665 & 0.7085 & 0.7165 & 0.7211 & 0.6749 \\
    \bottomrule
  \end{tabular}
  }

  \label{tab:layer_Ausburg_time100}
\end{table*}

\subsubsection{Performance as a function of diffusion model layers at time-steps 0, 50 and 100}
Quantitatively, as shown in \cref{tab:layer_Augsburg_time0}, \cref{tab:layer_Augsburg_timestep50} and \cref{tab:layer_Ausburg_time100}, the analysis of U-Net layer performance in the diffusion model for the Augsburg dataset across timesteps 0, 50, and 100 demonstrates a clear trend: both overall and per-class performance metrics improve for higher layers, peaking around Layer 10/layer 11.  Per class metrics all show significant gains at higher layers, indicating enhanced classification precision and agreement. 
\begin{figure}[h]
    \centering
    \vskip 0.15in

    \begin{minipage}[b]{0.13\textwidth}
        \includegraphics[height=0.4\textheight]{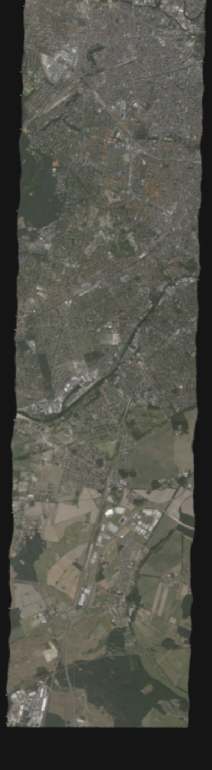}
        \centering \small \textbf{(a)} RGB Image
    \end{minipage}
    \hspace{0.05cm}
    \begin{minipage}[b]{0.13\textwidth}
        \includegraphics[height=0.4\textheight]{figure_Berlin/train_label_Berlin.png}
        \centering \small \textbf{(b)} Train Label
    \end{minipage}
    \hspace{0.05cm}
    \begin{minipage}[b]{0.13\textwidth}
        \includegraphics[height=0.4\textheight]{figure_Berlin/test_label_Berlin.png}
        \centering \small \textbf{(c)} Test Label
    \end{minipage}
    \hspace{0.05cm}
    \begin{minipage}[b]{0.13\textwidth}
        \includegraphics[height=0.4\textheight]{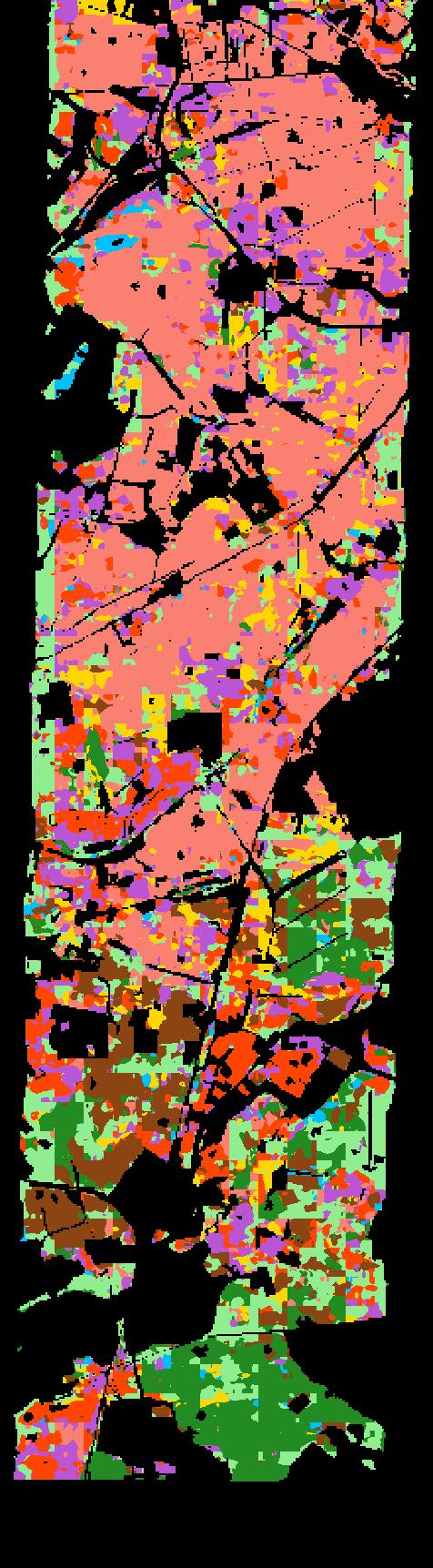}
        \centering \small \textbf{(d)} Layer 2
    \end{minipage}
    \hspace{0.05cm}
    \begin{minipage}[b]{0.13\textwidth}
        \includegraphics[height=0.4\textheight]{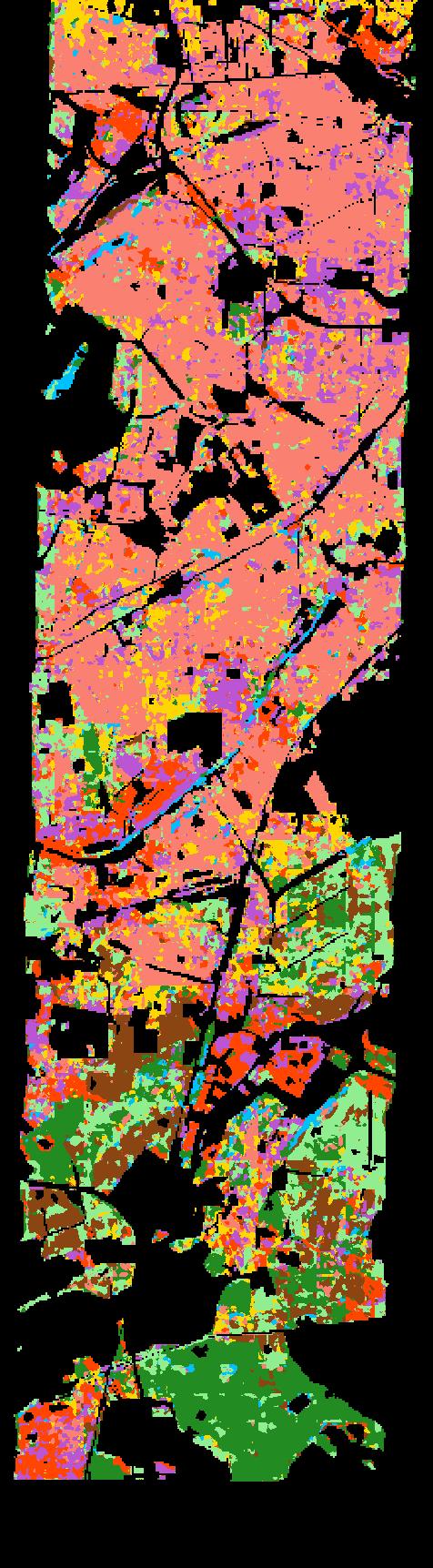}
        \centering \small \textbf{(e)} Layer 5
    \end{minipage}
    \hspace{0.05cm}
    \begin{minipage}[b]{0.13\textwidth}
        \includegraphics[height=0.4\textheight]{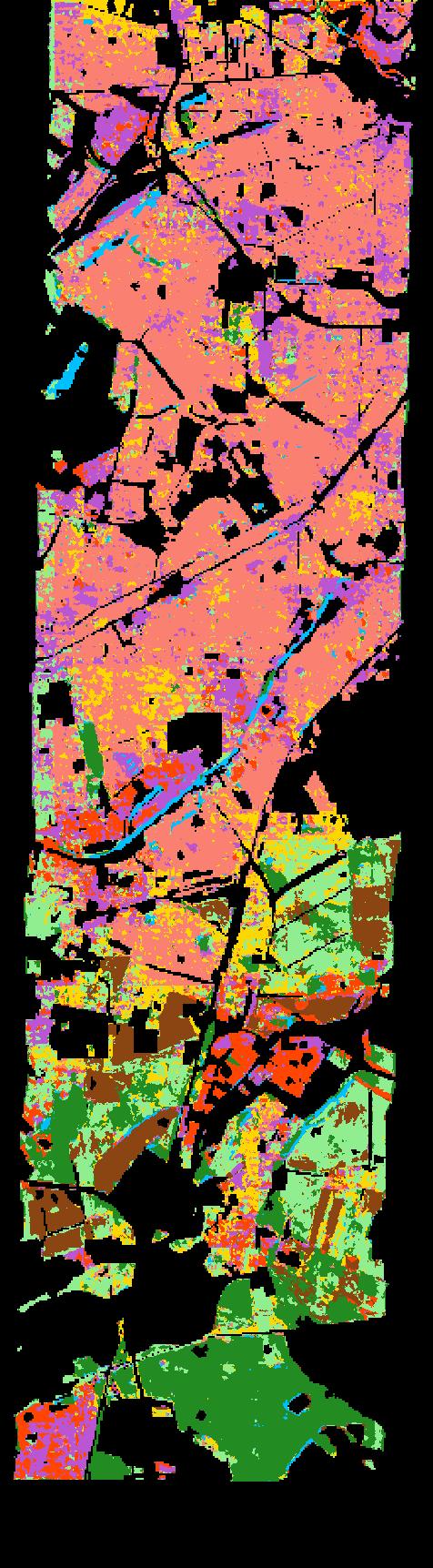}
        \centering \small \textbf{(f)} Layer 8
    \end{minipage}
    \hspace{0.05cm}
    \begin{minipage}[b]{0.13\textwidth}
        \fcolorbox{red}{white}{\includegraphics[height=0.4\textheight]{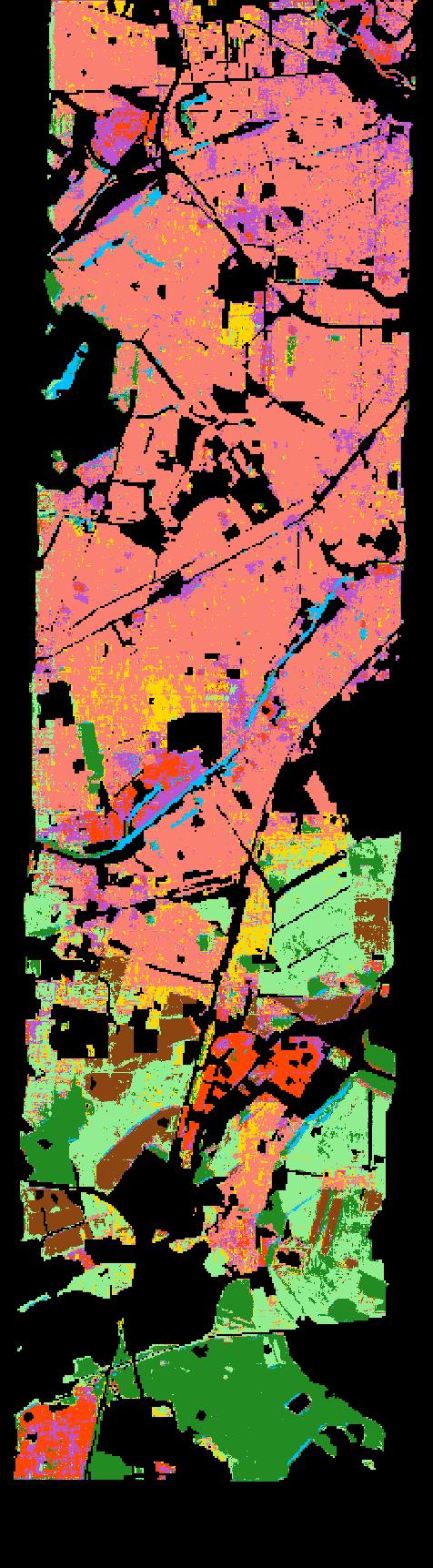}}
        \centering \small \textbf{(g)} Layer 11
    \end{minipage}

    \vskip 0.6em
    \includegraphics[width=\textwidth]{figure_Berlin/legend_horizontal.pdf}

    \caption{Visualization of informative layers from GeoDiffNet with a fixed timestep of 50 for the Berlin dataset.
    (a) RGB bands from HSI data, (b) training labels, (c) test labels for comparison.
    (d)–(g): Outputs from Layer 2, 5, 8, and 11 of GeoDiffNet, with Layer 11 yielding the clearest spatial features.}
    \label{fig:Berlin_layer2-5-8-11}
    \vskip -0.15in
\end{figure}

\begin{table}[h]
  \caption{Performance metrics across decoder layers in the diffusion model at \textbf{timestep 0} on the \textbf{Berlin} dataset.}
  \label{tab:layer_Berlin_timestep0}
  \centering
  \scriptsize
  \setlength{\tabcolsep}{3pt}
  \renewcommand{\arraystretch}{0.95}
  \resizebox{\textwidth}{!}{%
  \begin{tabular}{@{}lcccccccccccc@{}}
    \toprule
    & \multicolumn{12}{c}{Decoder layer} \\
    \cmidrule(lr){2-13}
    & 1 & 2 & 3 & 4 & 5 & 6 & 7 & 8 & 9 & 10 & 11 & 12 \\
    \midrule
    \textbf{Forest}            & 38.56 & 52.75 & 64.47 & 55.10 & 58.41 & 65.10 & 66.25 & 69.62 & 64.92 & 68.34 & 63.12 & 62.90 \\
    \textbf{Residential Area}  & 59.11 & 53.90 & 50.57 & 64.03 & 62.27 & 68.87 & 66.76 & 57.65 & 64.44 & 66.86 & 60.65 & 43.53 \\
    \textbf{Industrial Area}   & 52.76 & 50.15 & 53.30 & 47.72 & 50.50 & 51.01 & 58.71 & 46.42 & 42.80 & 31.76 & 45.54 & 42.29 \\
    \textbf{Low Plants}        & 30.74 & 23.27 & 32.32 & 35.19 & 47.49 & 51.82 & 50.42 & 56.65 & 62.45 & 61.25 & 73.47 & 72.55 \\
    \textbf{Soil}              & 58.19 & 69.31 & 75.92 & 71.57 & 68.91 & 81.53 & 70.89 & 80.98 & 72.42 & 60.24 & 81.88 & 72.24 \\
    \textbf{Allotment}         & 23.13 & 36.28 & 43.72 & 33.67 & 46.69 & 43.95 & 46.32 & 55.62 & 54.58 & 43.20 & 59.08 & 63.15 \\
    \textbf{Commercial Area}   & 38.93 & 43.87 & 49.03 & 38.02 & 37.31 & 33.12 & 42.19 & 38.40 & 38.04 & 36.92 & 41.76 & 42.94 \\
    \textbf{Water}             & 26.87 & 30.02 & 36.43 & 38.45 & 48.28 & 57.97 & 46.74 & 55.28 & 44.05 & 33.51 & 65.24 & 56.75 \\
    \midrule
    \textbf{Overall Accuracy (\%)} & 50.23 & 48.91 & 50.46 & 56.30 & 57.73 & 63.22 & 62.27 & 58.22 & 61.68 & 61.89 & 61.76 & 51.24 \\
    \textbf{Average Accuracy (\%)} & 41.04 & 44.94 & 50.72 & 47.97 & 52.48 & 56.67 & 56.03 & 57.58 & 55.46 & 50.26 & 61.34 & 57.05 \\
    \textbf{Kappa Coefficient}     & 0.3246 & 0.3275 & 0.3584 & 0.3964 & 0.4218 & 0.4813 & 0.4735 & 0.4405 & 0.4684 & 0.4604 & 0.4807 & 0.3854 \\
    \textbf{Mean IoU}              & 0.2275 & 0.2372 & 0.2659 & 0.2786 & 0.3097 & 0.3434 & 0.3469 & 0.3413 & 0.3412 & 0.3227 & 0.3669 & 0.3203 \\
    \textbf{Mean F1 Score}         & 0.3493 & 0.3642 & 0.4002 & 0.4126 & 0.4515 & 0.4855 & 0.4906 & 0.4838 & 0.4815 & 0.4555 & 0.5090 & 0.4619 \\
    \bottomrule
  \end{tabular}
  }
\end{table}

\begin{table}[ht]
  \caption{Performance metrics across decoder layers in the diffusion model at \textbf{timestep 50} on the \textbf{Berlin} dataset}
  \label{tab:layer_Berlin_timestep50}
  \centering
  \scriptsize
  \setlength{\tabcolsep}{3pt}
  \renewcommand{\arraystretch}{0.95}
  \resizebox{\textwidth}{!}{%
  \begin{tabular}{@{}lcccccccccccc@{}}
    \toprule
    & \multicolumn{12}{c}{Decoder layer} \\
    \cmidrule(lr){2-13}
    & 1 & 2 & 3 & 4 & 5 & 6 & 7 & 8 & 9 & 10 & 11 & 12 \\
    \midrule
    \textbf{Forest}            & 41.61 & 46.89 & 51.29 & 57.51 & 55.62 & 67.41 & 63.74 & 68.91 & 62.08 & 70.53 & \textbf{75.35} & 72.16 \\
    \textbf{Residential}       & 64.12 & 59.10 & 61.82 & 51.17 & 60.95 & 68.47 & 63.88 & 63.92 & 63.06 & 68.57 & \textbf{77.82} & 52.94 \\
    \textbf{Industrial}        & 46.45 & 51.75 & 50.74 & 36.78 & 43.42 & 47.00 & 46.77 & 43.74 & 24.27 & 40.98 & \textbf{53.41} & 47.86 \\
    \textbf{Low Plants}        & 22.13 & 31.78 & 26.54 & 31.97 & 37.30 & 39.35 & 35.64 & 46.82 & 61.76 & 53.11 & \textbf{71.50} & 62.40 \\
    \textbf{Soil}              & 52.64 & 71.20 & 68.97 & 70.70 & 67.28 & 70.45 & 73.14 & 74.44 & 61.76 & 79.67 & \textbf{75.12} & 65.28 \\
    \textbf{Allotment}         & 25.63 & 23.17 & 18.35 & 21.66 & 27.56 & 33.00 & 45.34 & 34.60 & 32.56 & 47.99 & \textbf{36.86} & 48.98 \\
    \textbf{Commercial}        & 25.30 & 33.70 & 32.83 & 45.36 & 31.65 & 39.32 & 45.81 & 46.20 & 61.74 & 45.94 & \textbf{40.41} & 64.45 \\
    \textbf{Water}             & 8.67  & 17.04 & 14.24 & 33.80 & 29.11 & 41.14 & 50.29 & 53.18 & 50.09 & 50.58 & \textbf{54.74} & 47.02 \\
    \midrule
    \textbf{Overall Accuracy (\%)}  & 51.02 & 51.37 & 52.45 & 48.21 & 53.87 & 60.89 & 58.23 & 59.97 & 60.02 & 64.06 & \textbf{72.15} & 57.08 \\
    \textbf{Average Accuracy (\%)}  & 35.82 & 41.83 & 40.60 & 43.62 & 44.11 & 50.77 & 53.08 & 53.98 & 52.17 & 57.17 & \textbf{60.65} & 57.63 \\
    \textbf{Kappa Coefficient}      & 0.3072 & 0.3358 & 0.3443 & 0.3228 & 0.3662 & 0.4438 & 0.4241 & 0.4462 & 0.4495 & 0.4917 & \textbf{0.5850} & 0.4360 \\
    \textbf{Mean IoU}               & 0.2054 & 0.2294 & 0.2331 & 0.2375 & 0.2522 & 0.3126 & 0.3102 & 0.3270 & 0.3341 & 0.3586 & \textbf{0.4218} & 0.3634 \\
    \textbf{Mean F1 Score}          & 0.3151 & 0.3509 & 0.3497 & 0.3632 & 0.3783 & 0.4487 & 0.4504 & 0.4677 & 0.4708 & 0.5025 & \textbf{0.5629} & 0.5077 \\
    \bottomrule
  \end{tabular}
  }
\end{table}

\begin{table}[h]
  \caption{Performance metrics across decoder layers in the diffusion model at\textbf{ timestep 100} on the \textbf{Berlin} dataset}
  \label{tab:layer_Berlin_timestep100}
  \centering
  \scriptsize
  \setlength{\tabcolsep}{3pt}
  \renewcommand{\arraystretch}{0.95}
  \resizebox{\textwidth}{!}{%
  \begin{tabular}{@{}lcccccccccccc@{}}
    \toprule
    & \multicolumn{12}{c}{Decoder layer} \\
    \cmidrule(lr){2-13}
    & 1 & 2 & 3 & 4 & 5 & 6 & 7 & 8 & 9 & 10 & 11 & 12 \\
    \midrule
    \textbf{Forest}            & 34.93 & 43.26 & 44.97 & 52.28 & 52.38 & 57.18 & 59.66 & 57.65 & 58.18 & 66.15 & 62.33 & 55.82 \\
    \textbf{Residential}       & 54.48 & 55.39 & 57.23 & 57.96 & 57.43 & 63.49 & 53.49 & 59.29 & 60.49 & 60.29 & 52.80 & 46.99 \\
    \textbf{Industrial}        & 42.27 & 32.49 & 28.64 & 18.05 & 30.75 & 37.95 & 38.93 & 41.17 & 36.93 & 39.07 & 40.88 & 40.72 \\
    \textbf{Low Plants}        & 18.32 & 19.48 & 17.04 & 24.92 & 25.01 & 29.01 & 28.15 & 42.90 & 46.14 & 46.99 & 57.17 & 61.16 \\
    \textbf{Soil}              & 56.90 & 44.43 & 66.18 & 65.56 & 74.41 & 75.75 & 79.67 & 73.89 & 74.60 & 81.11 & 75.01 & 74.26 \\
    \textbf{Allotment}         & 16.05 & 18.63 & 19.60 & 18.60 & 14.51 & 16.43 & 37.44 & 37.69 & 42.66 & 38.87 & 64.45 & 51.58 \\
    \textbf{Commercial}        & 26.39 & 36.43 & 44.55 & 47.10 & 49.73 & 51.25 & 53.76 & 55.04 & 52.16 & 51.43 & 47.82 & 65.08 \\
    \textbf{Water}             & 5.40  & 10.72 & 7.37  & 22.15 & 31.01 & 25.90 & 36.51 & 45.63 & 46.82 & 43.76 & 45.32 & 52.55 \\
    \midrule
    \textbf{Overall Accuracy (\%)}  & 43.88 & 45.36 & 47.37 & 49.52 & 50.23 & 55.24 & 50.68 & 55.77 & 56.80 & 57.87 & 54.77 & 51.76 \\
    \textbf{Average Accuracy (\%)}  & 31.84 & 32.60 & 35.70 & 38.33 & 41.90 & 44.62 & 48.45 & 51.66 & 52.25 & 53.46 & 55.72 & 56.02 \\
    \textbf{Kappa Coefficient}      & 0.2368 & 0.2631 & 0.2820 & 0.3088 & 0.3210 & 0.3748 & 0.3496 & 0.3994 & 0.4106 & 0.4260 & 0.4092 & 0.3835 \\
    \textbf{Mean IoU}               & 0.1683 & 0.1831 & 0.2005 & 0.2126 & 0.2357 & 0.2605 & 0.2611 & 0.3080 & 0.3174 & 0.3209 & 0.3434 & 0.3137 \\
    \textbf{Mean F1 Score}          & 0.2662 & 0.2860 & 0.3086 & 0.3272 & 0.3576 & 0.3851 & 0.3929 & 0.4490 & 0.4581 & 0.4609 & 0.4897 & 0.4584 \\
    \bottomrule
  \end{tabular}
  }
\end{table}

\subsection{Berlin Dataset}
\subsubsection{Visualization}

Using the same method, we test on the Berlin dataset with a fixed timestep of 50 and evaluate layer informativeness based on performance. As shown in \cref{fig:Berlin_layer2-5-8-11}, it presents the different layers (Layers 2, 5, 8, and 11) from the GeoDiffNet model for the Berlin dataset, comparing with the test label. higher layers provide more detailed and accurate feature representations. Specifically, higher layers (Layer 11) demonstrate more refined and precise feature delineations and improved classification accuracy compared to the lower layers.

\subsubsection{Peformance over layer at timestep 0, 50, 100}

Quantitatively, as shown in \cref{tab:layer_Berlin_timestep0}, \cref{tab:layer_Berlin_timestep50}, and \cref{tab:layer_Berlin_timestep100}
, we conducted the same experiment for the Berlin dataset, analyzing different layers across timesteps 0, 50, and 100. The results demonstrate a consistent trend: both overall and per-class performance metrics improve with increasing layer depth, peaking around Layer 11. 

\section{The Impact of Time Steps(noise)}
\label{sec:appendixB}
\subsection{Initial timesteps has more transferbilty}

\( t = 0 \), we extract the feature representation of the clean image \( x_0 \). As \( t \) increases, more noise is added, transforming the image to \( x_t \).
In this ablation study, we investigate the impact of different timesteps on feature extraction in diffusion models.

There are varying perspectives on the optimal timestep selection, largely depending on the dataset. According to \cite{xu2023open}, clean images, devoid of noise, extract the most optimal features. Conversely, \cite{luo2023diffusion} posits that the choice of timestep acts as a control mechanism, determining the level of high-frequency detail retained in the images. This selection helps to implicitly map noisy inputs to smoother outputs, thereby enhancing classification accuracy and overall model performance.

\cite{zhong2024diffusion}, in their \emph{Chain of Forgetting theorem}, elucidates how a diffusion model manages the denoising (generation) process over time. As \( t \to 0 \), the model zeroes in on the closest sample in the training dataset, executing a general denoising process with higher transferability. However, as \( t \to T \), the model's output aligns with the mean of the training data distribution, necessitating domain adaptation.

Although the Chain of Forgetting theorem primarily addresses the generation process, \cite{luo2023diffusion} observes that inversion (forward) processes contain information analogous to the generation process at the same timestep.
Based on their insights, we infer that during the forward (inversion) process of feature extraction, initial stages exhibit higher transferability, although the exact optimal timestep remains uncertain.

\subsection{Experiment amd results}
\begin{figure}[t]
    \centering
    \includegraphics[width=0.52\linewidth]{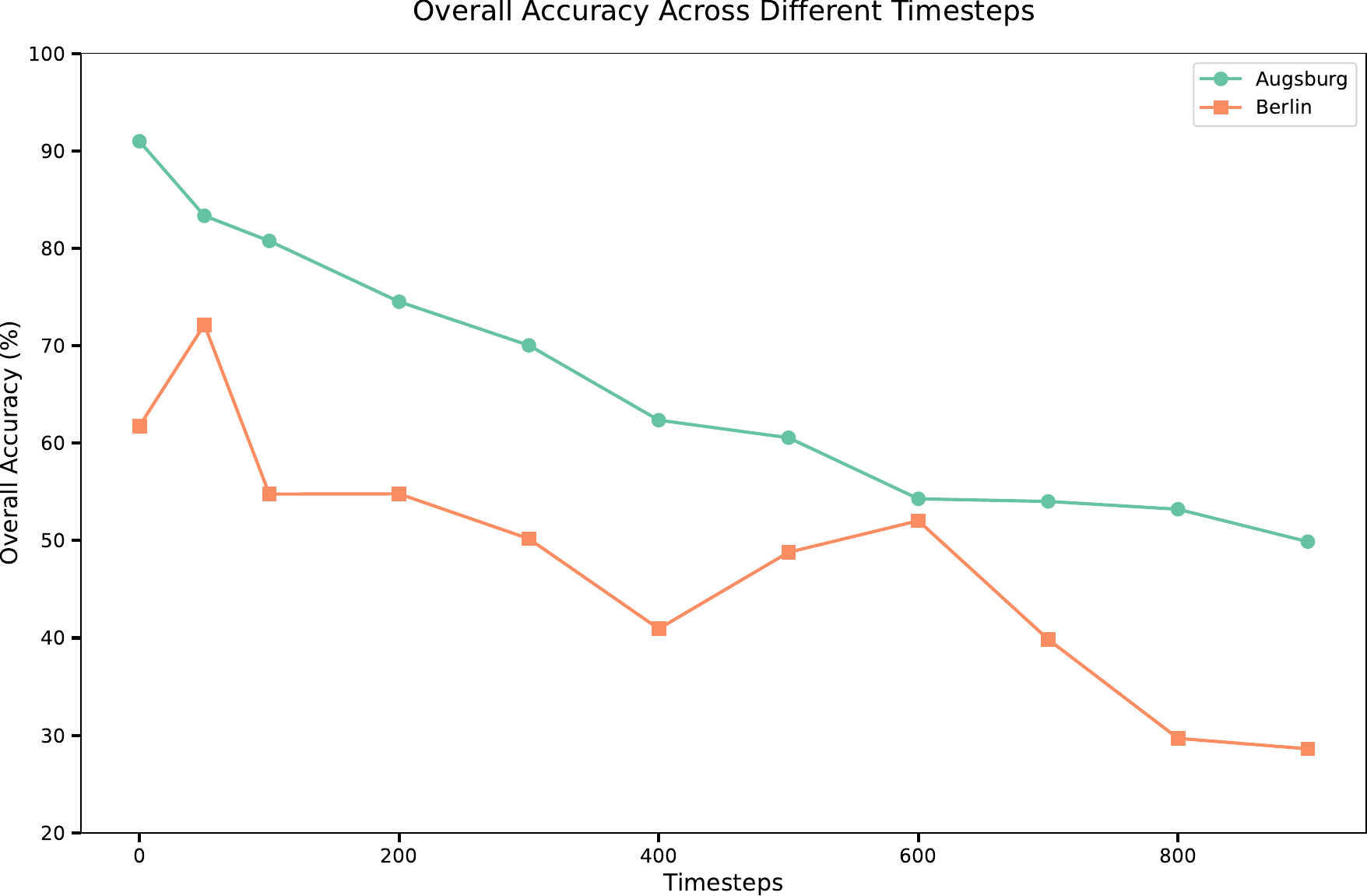}
    \caption{
    \small
    \textbf{Overall accuracy across diffusion timesteps $t$} for the Augsburg and Berlin datasets. 
    Accuracy is highest when features are extracted at early timesteps, indicating stronger transferability. 
    Optimal performance occurs at $t{=}0$ for Augsburg and $t{=}50$ for Berlin.
}
    \label{fig:oa_over_timesteps}

\end{figure}

We conducted experiments at initial timesteps (0, 50, 100) and continued with increments of 100 timesteps, with a fixed best layer: Layer 10 for the Augsburg dataset and Layer 11 for the Berlin dataset. Our objective was to evaluate the transferability of features extracted during the forward (inversion) process with different timesteps. The evaluation result can be seen from \cref{tab:Augsburg_timesep0-999_performance} and \cref{tab:timestep_0_999_performance_berlin}. 
From \cref{fig:oa_over_timesteps}, we can see that the best timesteps differ, with optimal performance observed at timestep 0 for Augsburg and timestep 50 for Berlin. However, the common observation is that performance decreases after the initial stages as noise increases, which matches our initial inference.

\begin{table}[h]
  \centering
   \caption{Performance metrics for GeoDiffNet at different \textbf{timesteps (0-900)} using \textbf{Layer 11} for spatial features extracted from the \textbf{Augsburg} dataset.}
  \setlength{\tabcolsep}{4pt}  
  \renewcommand{\arraystretch}{1.0}  
  {\small
  \begin{tabular}{@{}lcccccccccccc@{}}
    \toprule
    \textbf{Class} & \multicolumn{11}{c}{Timesteps} \\
    \cmidrule(lr){2-12}
    & TS 0 & TS 50 & TS 100 & TS 200 & TS 300 & TS 400 & TS 500 & TS 600 & TS 700 & TS 800 & TS 900 \\
    \midrule
    \textbf{Forest}            & 92.78 & 90.88 & 88.45 & 88.35 & 86.03 & 79.46 & 78.62 & 68.71 & 76.38 & 69.34 & 58.06 \\
    \textbf{Residential}       & 98.04 & 97.16 & 96.05 & 91.66 & 89.92 & 71.60 & 82.82 & 81.19 & 68.07 & 64.68 & 80.77 \\
    \textbf{Industrial}        & 61.46 & 25.56 & 26.61 & 14.70 & 10.68 & 5.85  & 7.08  & 8.02  & 1.15  & 0.99  & 1.20  \\
    \textbf{Low Plants}        & 95.98 & 81.85 & 77.09 & 66.20 & 57.05 & 59.72 & 42.10 & 29.75 & 41.82 & 46.89 & 24.66 \\
    \textbf{Allotment}         & 86.42 & 39.77 & 26.96 & 14.53 & 4.97  & 5.54  & 2.29  & 6.12  & 0.76  & 0.96  & 0.00  \\
    \textbf{Commercial}        & 6.35  & 4.33  & 3.72  & 3.36  & 3.48  & 0.73  & 1.65  & 0.49  & 0.24  & 0.00  & 0.12  \\
    \textbf{Water}             & 14.33 & 14.47 & 11.94 & 6.44  & 6.17  & 2.46  & 1.99  & 15.13 & 1.59  & 1.46  & 1.33  \\
    \midrule
    \textbf{Overall Accuracy (\%)}         & 90.98 & 83.33 & 80.75 & 74.51 & 70.03 & 62.35 & 60.55 & 54.28 & 54.01 & 53.21 & 49.88 \\
    \textbf{AA (\%)}                       & 65.05 & 50.58 & 47.26 & 40.75 & 36.90 & 32.20 & 30.94 & 29.92 & 27.15 & 26.33 & 23.73 \\
    \textbf{Kappa Coefficient}             & 0.8682 & 0.7549 & 0.7165 & 0.6216 & 0.5542 & 0.4414 & 0.4116 & 0.3359 & 0.3166 & 0.3037 & 0.2398 \\
    \textbf{Mean IoU}                      & 0.5482 & 0.4124 & 0.3857 & 0.3285 & 0.2883 & 0.2396 & 0.2290 & 0.2028 & 0.1912 & 0.1854 & 0.1584 \\
    \textbf{Mean F1 Score}                 & 0.6400 & 0.5092 & 0.4813 & 0.4228 & 0.3756 & 0.3211 & 0.3088 & 0.2869 & 0.2637 & 0.3023 & 0.2669 \\
    \bottomrule
  \end{tabular}
  }

  \label{tab:Augsburg_timesep0-999_performance}
\end{table}

\begin{table}[!h]
  \centering
  \setlength{\tabcolsep}{4pt}  
  \renewcommand{\arraystretch}{1.0}  
  \caption{Performance metrics for GeoDiffNet at different \textbf{timesteps (0-900)} using \textbf{Layer 10} for spatial features extracted from the \textbf{Berlin} dataset.}
  {\small
  \begin{tabular}{@{}lccccccccccc@{}}
    \toprule
    \textbf{Class} & \multicolumn{11}{c}{Timesteps} \\
    \cmidrule(lr){2-12}
    & TS 0 & TS 50 & TS 100 & TS 200 & TS 300 & TS 400 & TS 500 & TS 600 & TS 700 & TS 800 & TS 900 \\
    \midrule
    \textbf{Forest}            & 63.12 & 75.35 & 62.33 & 63.86 & 62.45 & 65.33 & 48.14 & 45.50 & 27.81 & 35.20 & 28.52 \\
    \textbf{Residential}       & 60.65 & 77.82 & 52.80 & 53.92 & 50.82 & 36.84 & 51.86 & 63.90 & 48.55 & 31.63 & 32.98 \\
    \textbf{Industrial}        & 45.54 & 53.41 & 40.88 & 34.64 & 50.94 & 49.57 & 59.36 & 64.58 & 50.61 & 33.82 & 18.55 \\
    \textbf{Low Plants}        & 73.47 & 71.50 & 57.17 & 59.39 & 45.71 & 29.62 & 47.40 & 23.84 & 16.94 & 16.56 & 18.83 \\
    \textbf{Soil}              & 81.88 & 75.12 & 75.01 & 70.45 & 64.22 & 74.76 & 74.21 & 81.00 & 68.87 & 59.61 & 39.60 \\
    \textbf{Allotment}         & 59.08 & 36.86 & 64.45 & 44.97 & 32.78 & 35.62 & 14.20 & 5.87 & 14.87 & 21.50 & 9.98 \\
    \textbf{Commercial}        & 41.76 & 40.41 & 47.82 & 41.67 & 31.89 & 35.39 & 22.76 & 10.98 & 19.53 & 14.51 & 21.11 \\
    \textbf{Water}             & 65.24 & 54.74 & 45.32 & 59.54 & 26.51 & 25.52 & 9.70 & 6.60 & 8.17 & 5.35 & 4.14 \\
    \midrule
    \textbf{Overall Accuracy (\%)}         & 61.76 & 72.15 & 54.77 & 54.79 & 50.18 & 40.94 & 48.79 & 52.03 & 39.85 & 29.69 & 28.62 \\
    \textbf{AA (\%)}                       & 61.34 & 60.65 & 55.72 & 53.56 & 45.66 & 44.08 & 40.95 & 37.78 & 31.92 & 27.27 & 21.71 \\
    \textbf{Kappa Coefficient}             & 0.4807 & 0.5850 & 0.4092 & 0.4009 & 0.3428 & 0.2689 & 0.3032 & 0.2960 & 0.1863 & 0.1234 & 0.0912 \\
    \textbf{Mean IoU}                      & 0.3669 & 0.4218 & 0.3434 & 0.3274 & 0.2930 & 0.2450 & 0.2331 & 0.2078 & 0.1595 & 0.1252 & 0.1028 \\
    \textbf{Mean F1 Score}                 & 0.5090 & 0.5629 & 0.4897 & 0.4694 & 0.4254 & 0.3228 & 0.3547 & 0.3143 & 0.2573 & 0.2110 & 0.1772 \\
    \bottomrule
  \end{tabular}
  }

  \label{tab:timestep_0_999_performance_berlin}
\end{table}

\section{Pre-trained Diffusion Model Architecture}
\label{sec:appendixD}

The pre-trained model uses OpenAI's pre-trained $64 \times 64$ diffusion model \cite{dhariwal2021diffusion}, which can find its detailed information and be downloaded from the following repository: 

\url{https://github.com/openai/guided-diffusion?tab=readme-ov-file}.

In the context of geospatial images, the $64 \times 64$ patch size provide extensive spatial context compared to conventional $11 \times 11$ patches used in fully-supervised HSI methods. Unlike traditional approaches that are constrained to small patches due to overfitting and training stability issues with limited supervision, our pre-trained diffusion backbone enables effective utilization of larger spatial contexts without data efficiency limitations.

\cref{tab:unet_decoder} lists the pre-trained model's decoder activation dimensions at different layers. Layers are numbered from bottom to top, including feature map resolution and channel dimension which facilitate extracting diffusion features at each layer. The design choices (i.e., the resolution and the number of channels) were determined experimentally.

\vspace*{0pt}  

\begin{table}[t]
  \centering
    \caption{Decoder architecture of the diffusion model, detailing image resolution, channels, and attention layers.}
  \setlength{\tabcolsep}{4pt}  
  \renewcommand{\arraystretch}{1.2}  
  \small{
  \begin{tabular}{@{}cccc@{}}
    \toprule
    \textbf{Layer} & \textbf{Resolution} & \textbf{Channels} & \textbf{Attention Layer} \\
    \midrule
    1  & $8 \times 8$   & 768 & \checkmark \\
    2  & $8 \times 8$   & 768 & \checkmark \\
    3  & $16 \times 16$ & 768 & \checkmark \\
    4  & $16 \times 16$ & 576 & \checkmark \\
    5  & $16 \times 16$ & 576 & \checkmark \\
    6  & $16 \times 16$ & 576 & \checkmark \\
    7  & $32 \times 32$ & 576 & \checkmark \\
    8  & $32 \times 32$ & 384 & \checkmark \\
    9  & $32 \times 32$ & 384 & \checkmark \\
    10 & $32 \times 32$ & 384 & \checkmark \\
    11 & $64 \times 64$ & 384 &  \\
    12 & $64 \times 64$ & 192 &  \\
    \bottomrule
  \end{tabular}
  }

  \label{tab:unet_decoder}
\end{table}

\vfill


\end{document}